\documentclass[aps,showpacs,twocolumn,groupedaddress]{revtex4-1}  % for review and submission
\usepackage{graphicx}  % needed for figures
\usepackage{dcolumn}   % needed for some tables
\usepackage{bm}    	% for math
\usepackage{amssymb}   % for math
\usepackage{amsmath}
\usepackage{url}
\urldef\mpfisingurl\url{https://github.com/Sohl-Dickstein/Minimum-Probability-Flow-Learning}

\usepackage{algorithm}
\usepackage{algorithmic}

\usepackage{verbatim}   % to comment out blocks of text

\newcommand{\set}[1]{\lbrace #1 \rbrace}

\newcommand{\dd}{{\mathrm{d}}}
\newcommand{\pd}[2]{\frac{\partial #1}{\partial #2}}

\newcommand{\avg}[1]{\left< #1 \right>}

\newcommand{\p}[2]{p_{#1}^{(#2)}}
\newcommand{\mb}{\mathbf}
\newcommand{\argmin}{\operatornamewithlimits{argmin}}
\newcommand{\doublebar}{\bigl|\!\bigr|}

\newcommand{\ham}[2]{\operatornamewithlimits{HAM}\left( #1 ; #2 \right)}

\begin{document}

% The following information is for internal review, please remove them for submission
\widetext

\title{A new method for parameter estimation in probabilistic models: Minimum probability flow}
\author{
Jascha Sohl-Dickstein$^{ad*}$, Peter Battaglino$^{bd*}$ and Michael R. DeWeese$^{bcd}$ \\
\small{$^a$Biophysics Graduate Group, $^{b}$Department of Physics, $^{c}$Helen Wills Neuroscience Institute} \\
\small{$^{d}$Redwood Center for Theoretical Neuroscience} \\
\small{University of California, Berkeley, 94720} \\
\small{\texttt{\{jascha, pbb, deweese\}@berkeley.edu}
\em$^*$These authors contributed equally.}
}

\begin{abstract}
Fitting probabilistic models to data is often difficult, due to the general intractability of the partition function. We propose a new parameter fitting method, Minimum Probability Flow (MPF), which is applicable to any parametric model. We demonstrate parameter estimation using MPF in two cases: a continuous state space model, and an Ising spin glass. In the latter case it outperforms current techniques by at least an order of magnitude in convergence time with lower error in the recovered coupling parameters.
\end{abstract}

\maketitle

Scientists and engineers increasingly confront large and
complex data sets that defy traditional modeling and analysis
techniques.
For example, fitting well-established probabilistic models from physics to population neural activity recorded in retina \cite{Schneidman_Nature_2006,Shlens_JN_2006,Schneidman_Nature_2006} or cortex \cite{Tang:2008p13129,Marre:2009p13182,Yu:2008p4351} is currently impractical for populations of more than about 100 neurons \cite{Broderick:2007p2761}.  
Similar difficulties occur in many other fields, including computer science \cite{mackay:book}, genomics \cite{genomics_review_2009}, and physics \cite{physics_param_est_book_2005}.

The canonical problem is to find parameter values $\theta$ that result in the
best match between a model and a list of observations $\mathcal{D}$. 
Parameter estimation can be viewed as the inverse of the usual problem
physicists face: 
rather than assuming fixed model parameters, such as coupling strengths in an Ising spin glass, and then predicting observable properties of the system, such as spin-spin correlations, our goal is to start with a series of observations and then estimate the underlying model parameters.
This is a challenging problem in many interesting cases~\cite{physics_param_est_book_2005,haykin2008nnc,mackay:book}.

Consider a data distribution over $N$ discrete states, represented as a vector $\mathbf{p}^{(0)}\in \mathbb R^N$, with $p^{(0)}_i$ the fraction of the observations $\mathcal{D}$ in state $i$.
A model distribution parameterized by $\theta$ is similarly represented as $\mb p^{(\infty)}\left( \theta \right)\in \mathbb R^N$.  
The superscripts $(0)$ and $(\infty)$ indicate initial conditions and equilibrium under system dynamics, as described below.  
For any model distribution $\mb p^{(\infty)}\left( \theta \right)$, the 
probability assigned to each state can be written
\begin{align}
\label{eqn:p infinity}
p^{(\infty)}_i\left( \theta \right) &= \frac
    	{\exp \left( -E_i\left( \theta \right) \right) }
    	{Z\left(\theta\right)}
		,
\end{align}
where $\mathbf{E}\left( \theta \right) \in \mathbb R^N$ can be viewed as the energy in the familiar Boltzmann distribution (with $k_B T$ set to 1).   $Z\left(\theta\right)$ is the partition function, 
which involves a sum over all $N$ possible states of the system,
\begin{align}
Z\left(\theta\right) &= \sum_i^N \exp \left( -E_i\left( \theta \right) \right)
.
\end{align}
For clarity we develop our method using discrete state spaces, but it extends to probabilistic models over continuous state spaces, as we demonstrate for an Independent Components Analysis (ICA) model~\cite{ICA}.

The standard objective for parameter estimation
is to maximize the likelihood of the model $\mathbf p^{(\infty)}\left( \theta \right)$ given the observations $\mathcal D$, or equivalently to minimize $D_{KL}\left( \mathbf p^{(0)} \doublebar  \mathbf p^{(\infty)}\left( \theta \right) \right)$, the KL divergence between the data distribution and model distribution~\cite{cover_thomas,mackay:book}.  The estimated parameters $\hat\theta$ are given by
\begin{align}
\hat \theta & = \argmin_\theta D_{KL}\left( \mathbf p^{(0)} \doublebar  \mathbf p^{(\infty)}\left( \theta \right) \right), \label{eq:KL min form}
\end{align}
\begin{align}
\label{eq:KL inf obj}
D_{KL}\left( \mathbf p^{(0)} \doublebar  \mathbf p^{(\infty)}\left( \theta \right) \right) & = \sum_i p_i^{(0)} \log p_i^{(0)}\\
& \qquad - \sum_i p_i^{(0)} \log p_i^{(\infty)} \left( \theta \right)  \nonumber
.
\end{align}
Unfortunately, the partition function $Z\left( \theta \right)$ in $p_i^{(\infty)} \left( \theta \right)$ usually involves an intractable sum over all system states.  This is commonly the major impediment to parameter estimation.

Many approaches exist for approximate parameter estimation, 
including mean field theory \cite{Pathria:1972p5861, Tanaka:1998p1984} and
its expansions \cite{Tanaka:1998p1984,hertz_spin_glass}, variational Bayes techniques \cite{Attias_bayes,Jaakkola_Jordan_bayes}, pseudolikelihood \cite{besag}, contrastive divergence \cite{Hinton02, Ackley85}, score matching \cite{Hyvarinen05,siwei2009}, minimum velocity learning~\cite{Movellan:2008p7643} and a multitude of Monte Carlo and numerical
integration-based methods~\cite{haykin2008nnc,Neal:HMC}.

Most Monte Carlo methods rely on two core concepts from statistical physics, which we will use to develop our parameter estimation technique, Minimum Probability Flow (MPF).
The first of these is conservation of probability, as enforced by the master equation for the evolution of a distribution $\mathbf p^{(t)}$ with time
\begin{align}
\label{eq:mastereqn}
\dot{p}_i^{(t)} &= \sum_{j\neq i} \Gamma_{ij}(\theta)\,p_j^{(t)} - \sum_{j\neq i} \Gamma_{ji}(\theta)\, p_i^{(t)}
.
\end{align}
$\Gamma_{ij}(\theta)$ gives the rate at which probability flows from
state $j$ into state $i$.
The first term of Eq.~\ref{eq:mastereqn} captures the flow of probability out of other states $j$ into the state $i$, and the second represents
the flow out of $i$ into other states $j$. 

The second core concept is detailed balance,
\begin{align}
\label{eq:detailed_balance}
\Gamma_{ji}\ p^{(\infty)}_i\left(\theta\right) &= \Gamma_{ij}\ p^{(\infty)}_j\left(\theta\right)
,
\end{align}
which when satisfied ensures that the probability flow from state $i$ into state $j$ equals the probability flow from $j$ into $i$, and thus that the distribution $\mathbf p^{(\infty)}$ is a fixed point of the dynamics.  Sampling in most Monte Carlo methods is performed by choosing $\mathbf \Gamma$ consistent with Eq. \ref{eq:detailed_balance} (and the added requirement of ergodicity \cite{mackay:book}), then stochastically running the dynamics in Eq. \ref{eq:mastereqn} .
Unfortunately, sampling algorithms can be exceedingly slow to converge, and are thus ill-suited for use in each parameter update step during parameter estimation.

\begin{figure}
\begin{center}
\parbox[c]{1.0\linewidth}{
\includegraphics[width=1.0\linewidth]{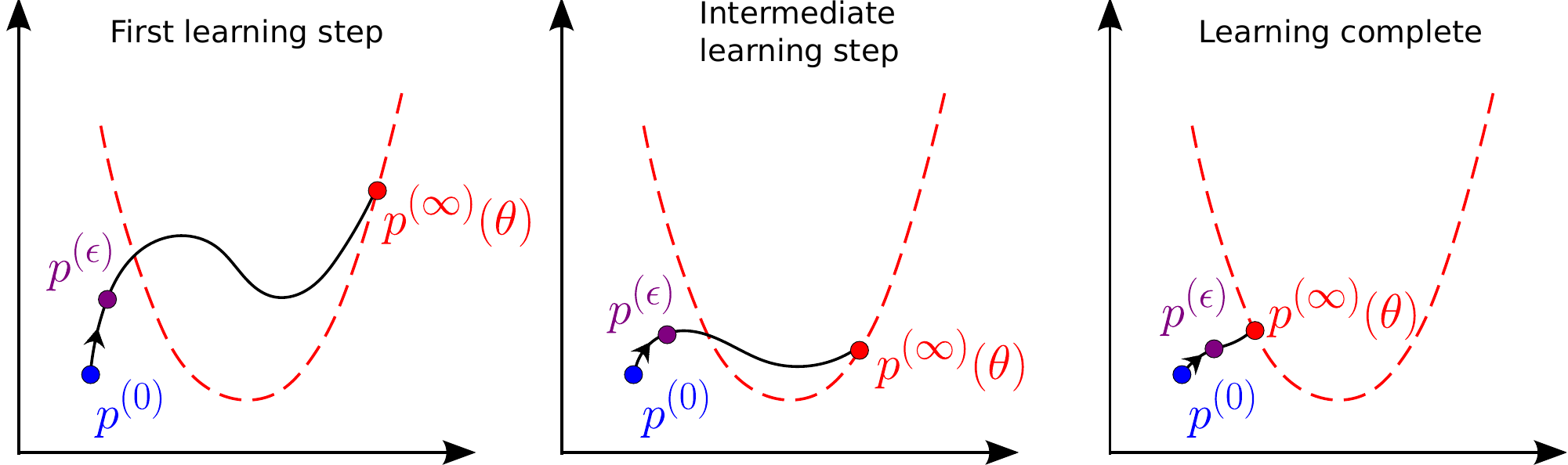}
} \\ \vspace{0.15in} \textsc{Progression of Learning} \vspace{-0.15in}
\end{center}
\caption{
An illustration of parameter estimation using Minimum Probability Flow (MPF). The three successive panels illustrate the sequence of parameter updates that occur during learning. Each set of axes represents the space of probability distributions. The dashed red curves indicate the family of model distributions $\mb{p}^{(\infty)}(\theta)$ parametrized by $\theta$.  The black curves indicate deterministic dynamics that transform the data distribution $\mb{p}^{(0)}$ into the model distribution $\mb{p}^{(\infty)}(\theta)$.  Under maximum likelihood learning, model
parameters $\theta$ are chosen so as to minimize the Kullback--Leibler (KL) divergence
between the data distribution $\mb{p}^{(0)}$ and the model distribution
$\mb{p}^{(\infty)}(\theta)$. Under MPF the KL divergence
between $\mb{p}^{(0)}$ and $\mb{p}^{(\epsilon)}(\theta)$ is minimized instead, where
$\mb{p}^{(\epsilon)}(\theta)$ is the distribution obtained by initializing the dynamics at the data distribution $\mb{p}^{(0)}$ and then evolving them for
an infinitesimal time $\epsilon$. Here we represent graphically how parameter updates that pull
$\mb{p}^{(\epsilon)}(\theta)$ towards $\mb{p}^{(0)}$ also tend to pull
$\mb{p}^{(\infty)}(\theta)$ towards $\mb{p}^{(0)}$.
}
\label{fig:KL}
\end{figure}

Using these two core concepts, we propose an approach, illustrated in Fig.~\ref{fig:KL}, that avoids both sampling and explicit calculation of the partition function.
Specifically, we establish deterministic dynamics obeying Eqs. \ref{eq:mastereqn} and \ref{eq:detailed_balance} that converge to the model distribution $\mathbf{p}^{(\infty)}\left( \theta \right)$, and initialize them at the data distribution $\mathbf{p}^{(0)}$.  Rather than
allowing the dynamics to fully converge
and making parameter updates that minimize $D_{KL}\left( \mathbf p^{(0)} \doublebar  \mathbf  p^{(\infty)}\left( \theta \right) \right)$ as in maximum likelihood learning (Eqs.~\ref{eq:KL min form},\ref{eq:KL inf obj}), our parameter updates instead minimize $D_{KL}\left( \mathbf p^{(0)} \doublebar  \mathbf p^{(\epsilon)}\left( \theta \right) \right)$, the KL divergence after running the dynamics for an infinitesimal time $\epsilon$.  This requires computing only the instantaneous flow of probability away from the data distribution at time $t=0$.

The transition rates $\Gamma_{ij}$ are under-constrained by Eq. \ref{eq:detailed_balance}.
Introducing the additional constraint that $\mathbf{\Gamma}$ be invariant to the addition of a constant to the energy function (as is true for the model distribution $\mb{p}^{(\infty)}(\theta)$),
we choose the following form for $\Gamma_{ij}$:
\begin{align}
\label{eqn:gamma symmetric}
\Gamma_{ij} &=
	g_{ij}\exp \left[ \frac{1}{2} \left( E_j\left( \theta \right)-E_i\left(  \theta \right) \right)\right]\ \ \ \ \ \ \ \ \  \left(i \neq j\right),
\end{align}
where $g_{ij}=g_{ji} \in\{0,1\}$.  
The vast majority of the factors $g_{ij}$ can generally be set to 0.
However, for the dynamics to converge to $\mathbf p^{(\infty)}\left( \theta \right)$, there must be sufficient non-zero $\mathbf\Gamma$ elements to ensure mixing.
In binary systems, good results are obtained by setting $g_{ij}=g_{ji}=1$ only for states $i$ and $j$ differing by a single bit flip.   
The elements of $g_{ij}$ may also be sampled, rather than set by a deterministic scheme (see Appendix).

Given the transition matrix $\mathbf \Gamma$ and the list $\mathcal{D}$ of observed data samples, and taking $\epsilon$ small, the objective function $D_{KL}\left( \mathbf p^{(0)} \doublebar  \mathbf p^{(\epsilon)}\left( \theta \right) \right)$ is approximated by its first order Taylor expansion, denoted $K\left(\theta\right)$ (see Appendix)
\begin{align}
K\left( \theta \right) & = 
D_{KL}\left(
\mathbf{p^{(0)}} \doublebar \mathbf{p^{(t)}}
\left(\theta\right)\right)\Big |_{t=0} \\
& \qquad + \epsilon \frac
{\partial D_{KL}\left(
\mathbf{p^{(0)}} \doublebar \mathbf{p^{(t)}}
\left(\theta\right)\right)}
{\partial t}\Big |_{t=0} \nonumber
\\
& = \epsilon \sum_{i\notin \mathrm{\mathcal{D}}} \dot{p}_i^{(0)}
=  \frac{\epsilon}{M}\sum_{i\notin \mathrm{\mathcal{D}}}\sum_{j\in \mathrm{\mathcal{D}}} \Gamma_{ij} \label{eqn:flow} \\
& =  \frac{\epsilon}{M} \sum_{j\in \mathrm{\mathcal{D}}
} \sum_{
 	i\notin \mathrm{\mathcal{D}} 
}
g_{ij} \exp \left[ \frac{1}{2} \left( E_j\left( \theta \right)-E_i\left(  \theta \right) \right)\right]
\label{eq:K_finalform}
,
\end{align}
where $M=\left|\mathcal{D}\right|$ is the number of data samples.  
Parameter estimation is performed by finding
$\hat \theta  = \argmin_\theta K\left( \theta \right)$
generally via gradient descent of $K\left( \theta \right)$.
Thus, minimizing the KL divergence $D_{KL}\left( p^{(0)} \doublebar  p^{(\epsilon)}\left( \theta \right) \right)$ for small $\epsilon$ is equivalent to minimizing the initial flow of probability out of data-states $j$ into non-data states $i$ (Eq.~\ref{eqn:flow}).  For small systems, or large numbers of observations, every state may be a data state, in which case the first order term vanishes and higher order terms must be included.

The dimensionalities of $\mathbf{p}^{(0)}$ and $\mathbf{\Gamma}$ are typically large ({\it{e.g.}}, $2^d$ and $2^d \times 2^d$, respectively, for a $d$-bit binary
system).  Fortunately, both can also be made extremely sparse:   $p_j^{(0)} = 0$ for all non-data states $j \notin \mathcal D$, and we need only evaluate $\Gamma_{ij}$ for which $j\in\mathcal{D}$ and $g_{ij}\neq0$.  
The cost in both memory and time is therefore only $\mathcal{O}(M)$ per learning step.  The dependence of total convergence time on the number of samples $M$ is problem dependent, but it is roughly $\mathcal{O}(M)$ 
for the Ising spin glass model discussed below (see Appendix).

In addition, when estimating parameters for a model in the exponential family --- that is, models such as spin glasses for which the energy function is linear in the parameters $\theta$ --- the MPF objective function $K\left(\theta\right)$ is convex \cite{macke}, guaranteeing that the global minimum can always be found via gradient descent.  For exponential family models over continuous rather than discrete state spaces, MPF further provides a closed form solution for parameter estimation \cite{hyvarinen2007}.  MPF is also consistent --- meaning that if the data distribution $\mathbf{p}^{(0)}$ belongs to the family of distributions $\mathbf{p}^{(\infty)}\left( \theta \right)$ parameterized by $\theta$ (Fig. \ref{fig:KL}, red dashed line), the objective function $D_{KL}\left( p^{(0)} \doublebar  p^{(\epsilon)}\left( \theta \right) \right)$  will have its global minimum at the true parameter values (see Appendix).

We evaluated performance by fitting an Ising spin glass (sometimes referred to in the computer science literature as a fully visible Boltzmann machine or simply as an Ising model) of the form
\begin{align}
p^{(\infty)}(\mb{x};\mb{J}) &= \frac{1}{Z(\mb{J})}\exp\left[ -\mb{x}^{\mathrm{T}}\mb{J}\mb{x} \right],
\end{align}
where $\mb{J}$ only had non-zero elements corresponding to
nearest-neighbor units in a two-dimensional square lattice, and bias terms along the diagonal. The training data $\mathcal{D}$ consisted of % *********a list of
$M$ $d$-element iid binary samples $\mb x \in \{0,1\}^d$ generated via 
Swendsen-Wang sampling~\cite{swendsen1987nonuniversal} from a spin glass with known coupling parameters. In this example, we used a square $10 \times 10$ lattice, $d=10^2$.
The non-diagonal nearest-neighbor elements of $\mb{J}$ were set using draws from a normal distribution with variance $\sigma^2 = 10$.
The diagonal (bias) elements of $\mb{J}$ were set so that each column of $\mb{J}$ summed to 0, and the expected unit activations were $0.5$.
The $2^d \times 2^d$ element transition matrix $\mb{\Gamma}$ was populated sparsely, 
\begin{align}
\label{eqn:g_ij_ising}
g_{ij} = g_{ji} &=
\left\{\begin{array}{ccc}
	1 &  & \mathrm{states\ }i\mathrm{\ and\ }j\mathrm{\ differ\ by\ single\ bit\ flip} \\
	0 &  & \mathrm{otherwise}
\end{array}\right.
.
\end{align}
Code implementing MPF is available \footnote{\mpfisingurl}.

\begin{figure}
\begin{center}
\begin{tabular}{ccc}
\begin{tabular}{c}\includegraphics[width=0.3\linewidth]{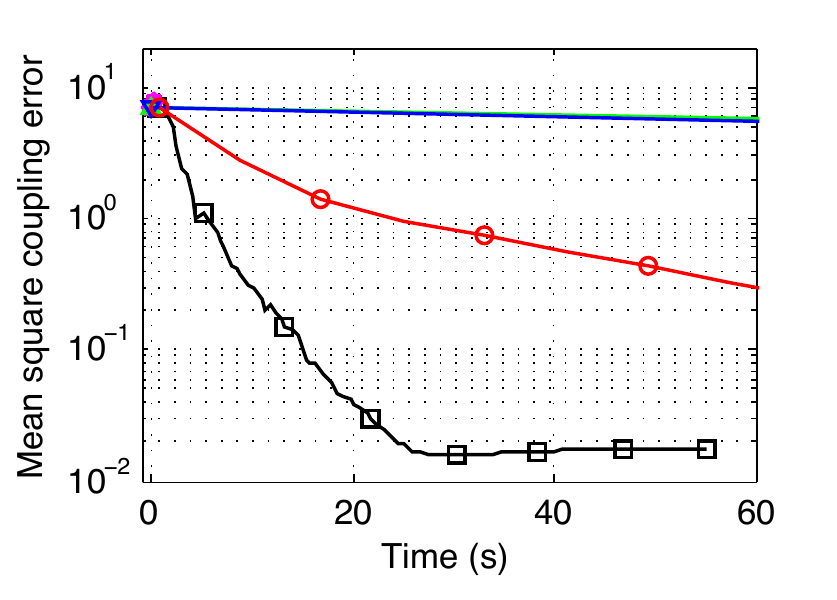}\vspace{-12pt}\\(a)\end{tabular}
&
\begin{tabular}{c}\includegraphics[width=0.3\linewidth]{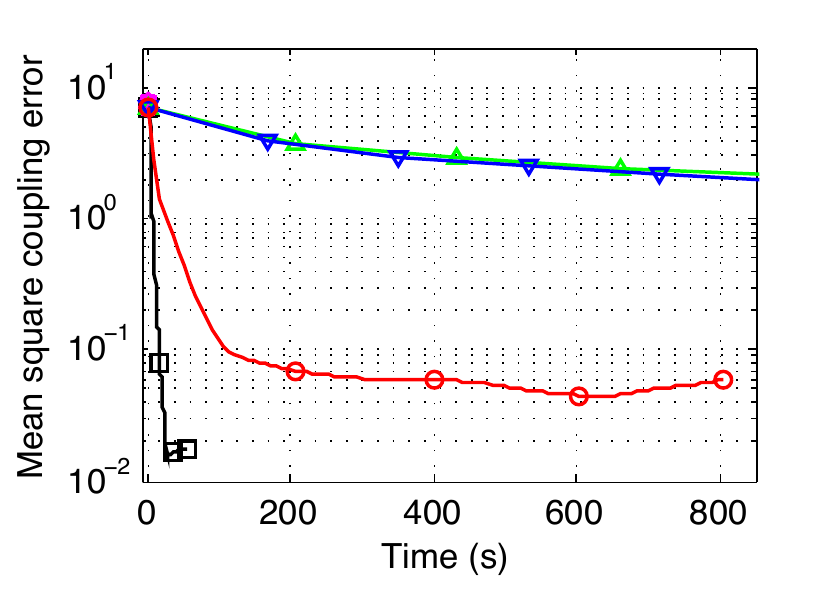}\vspace{-12pt}\\(b)\end{tabular}
&
\begin{tabular}{c}\includegraphics[width=0.3\linewidth]{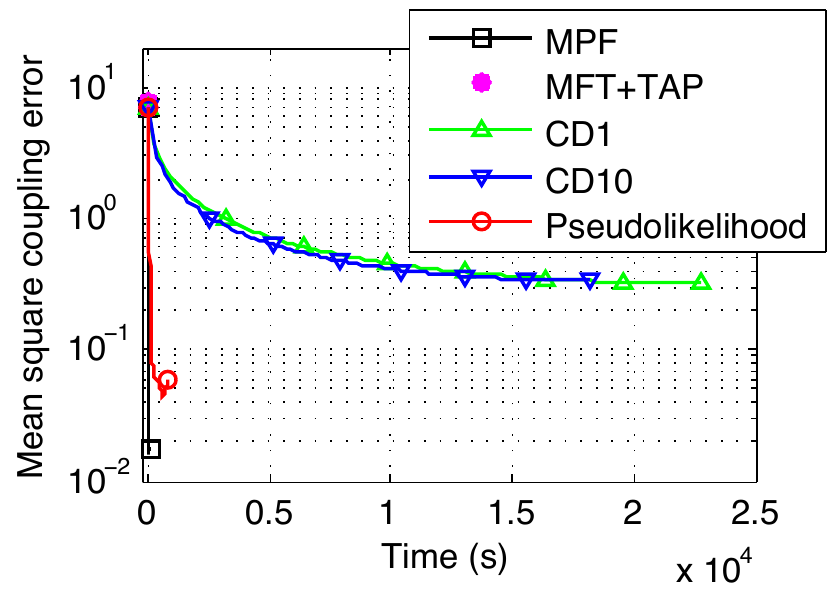}\vspace{-12pt}\\(c)\end{tabular}
\\
\begin{tabular}{c}\includegraphics[width=0.3\linewidth]{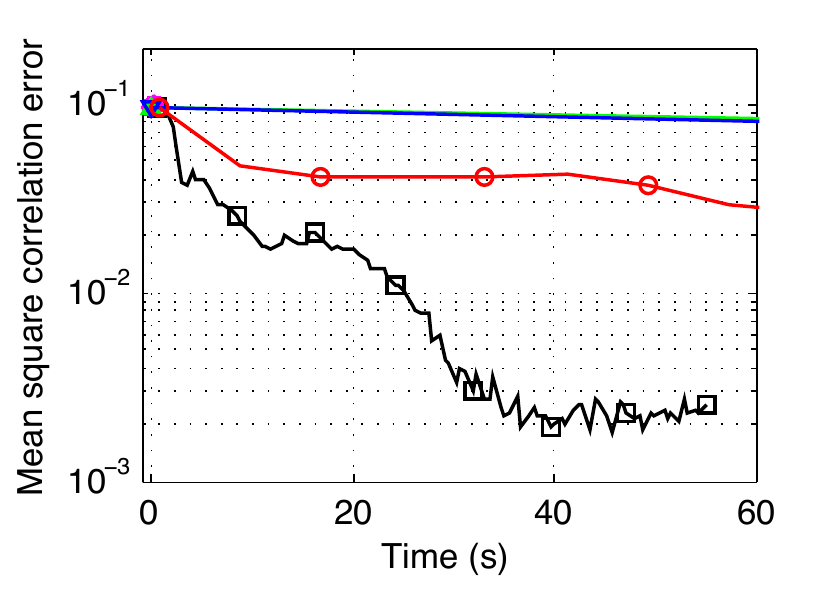}\vspace{-12pt}\\(d)\end{tabular}
&
\begin{tabular}{c}\includegraphics[width=0.3\linewidth]{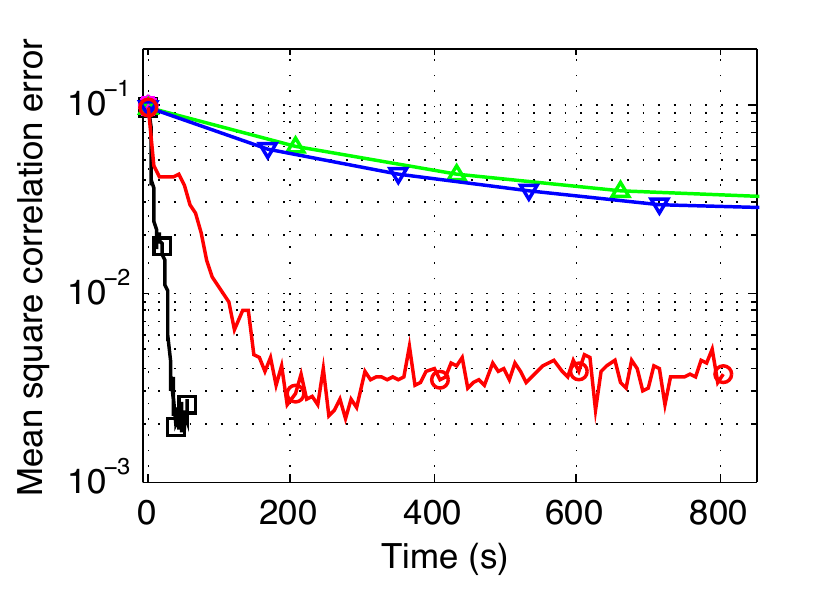}\vspace{-12pt}\\(e)\end{tabular}
&
\begin{tabular}{c}\includegraphics[width=0.3\linewidth]{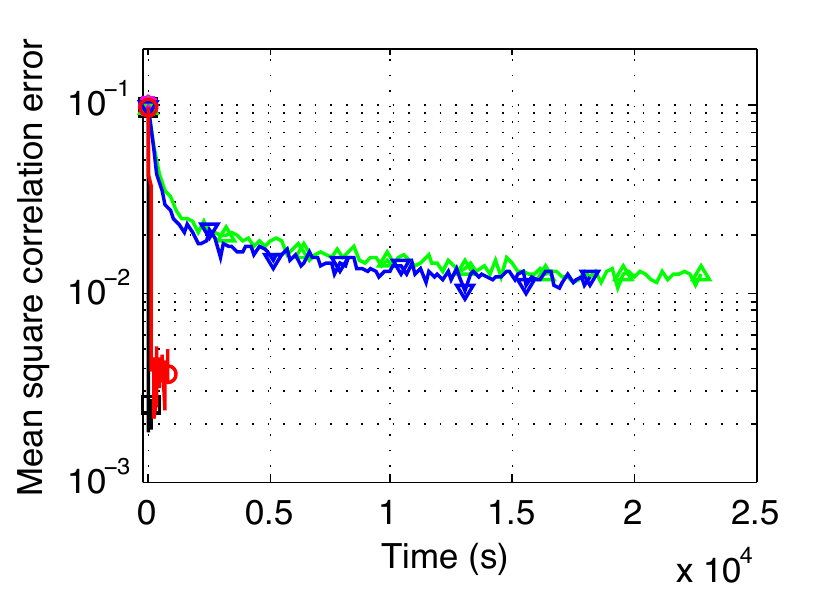}\vspace{-12pt}\\(f)\end{tabular}
\end{tabular}
\end{center}
\caption{A demonstration of Minimum Probability Flow (MPF) outperforming existing techniques for parameter recovery in an Ising spin glass model.
{\textbf{(a)}} Time evolution of the mean square error in the coupling strengths for 5 methods for the first 60 seconds of learning.  Note that mean field theory with second order corrections (MFT+TAP) actually increases the error above random parameter assignment, though it does converge to the correct answer in some other parameter regimes, such as in the high temperature limit of this Ising spin glass model~\cite{hertz_spin_glass}. {\textbf{(b)}} Mean square error in the coupling strengths for the first 800 seconds of learning. {\textbf{(c)}} Mean square error in coupling strengths for the entire learning period. {\textbf{(d)}}--{\textbf{(f)}} Mean square error in pairwise correlations for the first 60 seconds of learning, the first 800 seconds of learning, and the entire learning period, respectively. In every comparison above MPF finds a better fit, and for all cases but MFT+TAP does so in a shorter time.}
\label{fig:ising_compare}
\end{figure}

We compared parameter estimation using MPF against parameter estimation using four competing techniques: mean field theory (MFT) with Thouless-Anderson-Palmer (TAP) corrections \cite{tap_spin}, one-step and ten-step contrastive divergence \cite{Hinton02} 
(CD-1 and CD-10), and pseudolikelihood \cite{besag}.   The results of our simulations are shown in Fig.~\ref{fig:ising_compare}, which plots the mean square error in the recovered $\mb{J}$ and in the corresponding pairwise correlations as a function of learning time for MPF and the competing approaches outlined above.
Using MPF, learning took approximately 60 seconds, compared to roughly 800 seconds for pseudolikelihood and approximately 20,000 seconds for both 1-step and 10-step contrastive divergence. Reasonable steps were taken to optimize the performance of all the parameter estimation methods tested (see Appendix).   Note that, given sufficient samples, MPF is guaranteed to converge exactly to the right answer, as it is consistent and the objective function $K\left( \theta \right)$ is convex for a spin glass.   MPF fit the model to the data more accurately than any of the other techniques.  MPF was dramatically faster to converge than any of the other techniques tested, with the exception of MFT+TAP, which failed to find reasonable parameters. Note that MFT+TAP does converge to the correct answer in certain parameter regimes, such as the high temperature limit~\cite{hertz_spin_glass}, while remaining much faster than the other four techniques.  

\begin{figure}
\center{
%\framebox[0.5\textwidth]{
\parbox[b]{0.45\linewidth}{
\center{
\begin{tabular}{c}
\includegraphics[width= 0.9\linewidth]{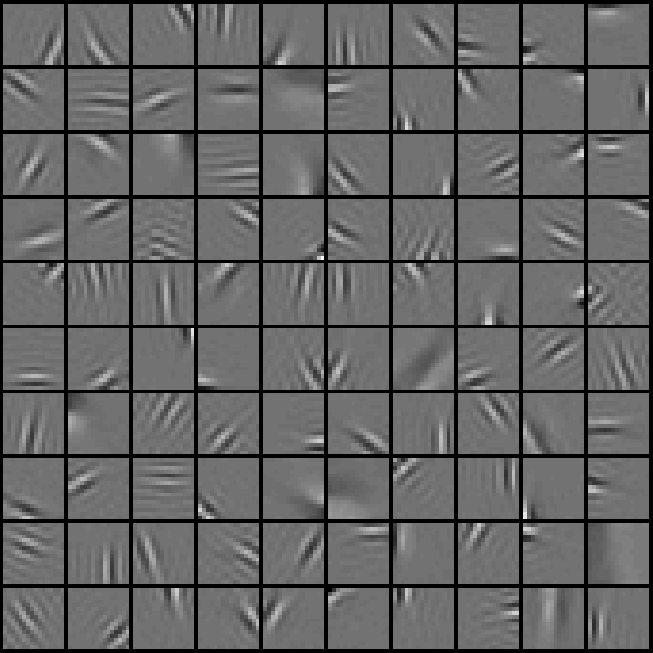} \\
(a)
\end{tabular}
}
}
\parbox[b]{0.45\linewidth}{
\center{
\begin{tabular}{c}
\includegraphics[width= 0.9\linewidth]{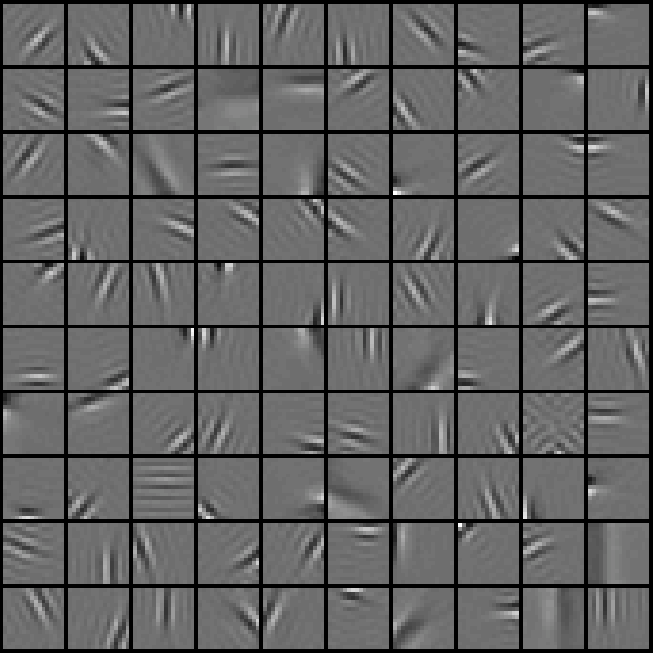} \\
(b)
\end{tabular}
}
}
%}
}
\caption{
A demonstration of Minimum Probability Flow (MPF) parameter estimation in a continuous state space probabilistic model.  Each square represents a $10 \times 10$ pixel filter $\mb J_k$ for an independent component analysis (ICA) model trained on natural image patches via {\textbf{(a)}} MPF or {\textbf{(b)}} maximum likelihood learning.
The visual similarity of the filters is consistent with the nearly identical average log likelihoods for the two models ($-120.61\ \mathrm{nats}$ and $-120.33\ \mathrm{nats}$ respectively).
}
\label{fig:ICA}
\end{figure}

As a demonstration of parameter estimation using MPF for a continuous state space probabilistic model, we trained the filters $\mb J \in \mathbb R^{d\times d}$ of a $d$ dimensional independent component analysis (ICA) \cite{ICA} model with a Laplace prior,
\begin{align}
p^{(\infty)}\left(\mb x; \mb J \right)
&=
\frac{
e^{
-\sum_k \left|
\mb J_{k} \mb x
\right|
}
}{
2^d
\det\left(\mb J^{-1} \right)
}
,
\end{align}
where $\mb x \in \mathbb R^d$ is a continuous state space.  Since the log likelihood and its gradient can be calculated analytically for ICA, we solved for $\mb J$ via maximum likelihood learning (Eq. \ref{eq:KL min form}) as well as MPF, and compared the resulting log likelihoods.  Training data consisted of natural image patches.  The log likelihood of the model trained with MPF was $-120.61\ \mathrm{nats}$, while that for the maximum likelihood trained model was a nearly identical $-120.33\ \mathrm{nats}$.  Average log likelihood at parameter initialization was $-189.23\ \mathrm{nats}$.  (see Appendix)  
The edge-like filters resulting from training, similar to receptive fields in the primary visual cortex \cite{dayan2001theoretical}, are shown in Fig. \ref{fig:ICA} for both maximum likelihood and MPF solutions.

In summary, we have presented a novel, general purpose framework, called Minimum Probability Flow (MPF), for fitting probabilistic models to data that outperforms current techniques in both learning time and accuracy. Our method works for any parametric model without hidden state variables, over either continuous or discrete state spaces, and we avoid explicit calculation of the partition function by employing deterministic dynamics in place of the slow sampling required by many existing approaches.
Because MPF provides a simple and well-defined objective function, it can be minimized quickly using existing higher order gradient descent techniques.
Furthermore, the objective function is convex for many models, including those in the exponential family, ensuring that the global minimum can be found with gradient descent.
Finally, MPF is consistent
--- it will find the true parameter values when the data distribution belongs to the same family of parametric models as the model distribution.

\subsubsection*{Acknowledgments}

We would like to thank Javier Movellan, Tamara Broderick, Miroslav Dud\'{\i}k, Ga\v{s}per Tka\v{c}ik, Robert E. Schapire, William Bialek for sharing work in progress and data; Ashvin Vishwanath, Jonathon Shlens, Tony Bell, Charles Cadieu, Nicole Carlson, Christopher Hillar, Kilian Koepsell, Bruno Olshausen and the rest of the Redwood Center for many useful discussions; and the James S. McDonnell Foundation (JSD, PB, JSD) and the Canadian Institute for Advanced Research - Neural Computation and Perception Program (JSD) for financial support.

\bibliography{prl} 

%merlin.mbs apsrev4-1.bst 2010-07-25 4.21a (PWD, AO, DPC) hacked
%Control: key (0)
%Control: author (8) initials jnrlst
%Control: editor formatted (1) identically to author
%Control: production of article title (-1) disabled
%Control: page (0) single
%Control: year (1) truncated
%Control: production of eprint (0) enabled
\begin{thebibliography}{40}%
\makeatletter
\providecommand \@ifxundefined [1]{%
 \@ifx{#1\undefined}
}%
\providecommand \@ifnum [1]{%
 \ifnum #1\expandafter \@firstoftwo
 \else \expandafter \@secondoftwo
 \fi
}%
\providecommand \@ifx [1]{%
 \ifx #1\expandafter \@firstoftwo
 \else \expandafter \@secondoftwo
 \fi
}%
\providecommand \natexlab [1]{#1}%
\providecommand \enquote  [1]{``#1''}%
\providecommand \bibnamefont  [1]{#1}%
\providecommand \bibfnamefont [1]{#1}%
\providecommand \citenamefont [1]{#1}%
\providecommand \href@noop [0]{\@secondoftwo}%
\providecommand \href [0]{\begingroup \@sanitize@url \@href}%
\providecommand \@href[1]{\@@startlink{#1}\@@href}%
\providecommand \@@href[1]{\endgroup#1\@@endlink}%
\providecommand \@sanitize@url [0]{\catcode `\\12\catcode `\$12\catcode
  `\&12\catcode `\#12\catcode `\^12\catcode `\_12\catcode `\%12\relax}%
\providecommand \@@startlink[1]{}%
\providecommand \@@endlink[0]{}%
\providecommand \url  [0]{\begingroup\@sanitize@url \@url }%
\providecommand \@url [1]{\endgroup\@href {#1}{\urlprefix }}%
\providecommand \urlprefix  [0]{URL }%
\providecommand \Eprint [0]{\href }%
\providecommand \doibase [0]{http://dx.doi.org/}%
\providecommand \selectlanguage [0]{\@gobble}%
\providecommand \bibinfo  [0]{\@secondoftwo}%
\providecommand \bibfield  [0]{\@secondoftwo}%
\providecommand \translation [1]{[#1]}%
\providecommand \BibitemOpen [0]{}%
\providecommand \bibitemStop [0]{}%
\providecommand \bibitemNoStop [0]{.\EOS\space}%
\providecommand \EOS [0]{\spacefactor3000\relax}%
\providecommand \BibitemShut  [1]{\csname bibitem#1\endcsname}%
\let\auto@bib@innerbib\@empty
%</preamble>
\bibitem [{\citenamefont {Schneidman}\ \emph {et~al.}(2006)\citenamefont
  {Schneidman}, \citenamefont {2nd}, \citenamefont {Segev},\ and\ \citenamefont
  {Bialek}}]{Schneidman_Nature_2006}%
  \BibitemOpen
  \bibfield  {author} {\bibinfo {author} {\bibfnamefont {E.}~\bibnamefont
  {Schneidman}}, \bibinfo {author} {\bibfnamefont {M.~J.~B.}\ \bibnamefont
  {2nd}}, \bibinfo {author} {\bibfnamefont {R.}~\bibnamefont {Segev}}, \ and\
  \bibinfo {author} {\bibfnamefont {W.}~\bibnamefont {Bialek}},\ }\href@noop {}
  {\bibfield  {journal} {\bibinfo  {journal} {Nature}\ }\textbf {\bibinfo
  {volume} {440}},\ \bibinfo {pages} {1007} (\bibinfo {year}
  {2006})}\BibitemShut {NoStop}%
\bibitem [{\citenamefont {Shlens}\ \emph {et~al.}(2006)\citenamefont {Shlens},
  \citenamefont {Field}, \citenamefont {Gauthier}, \citenamefont {Grivich},
  \citenamefont {Petrusca}, \citenamefont {Sher}, \citenamefont {Litke},\ and\
  \citenamefont {Chichilnisky}}]{Shlens_JN_2006}%
  \BibitemOpen
  \bibfield  {author} {\bibinfo {author} {\bibfnamefont {J.}~\bibnamefont
  {Shlens}}, \bibinfo {author} {\bibfnamefont {G.~D.}\ \bibnamefont {Field}},
  \bibinfo {author} {\bibfnamefont {J.~L.}\ \bibnamefont {Gauthier}}, \bibinfo
  {author} {\bibfnamefont {M.~I.}\ \bibnamefont {Grivich}}, \bibinfo {author}
  {\bibfnamefont {D.}~\bibnamefont {Petrusca}}, \bibinfo {author}
  {\bibfnamefont {A.}~\bibnamefont {Sher}}, \bibinfo {author} {\bibfnamefont
  {A.~M.}\ \bibnamefont {Litke}}, \ and\ \bibinfo {author} {\bibfnamefont
  {E.~J.}\ \bibnamefont {Chichilnisky}},\ }\href@noop {} {\bibfield  {journal}
  {\bibinfo  {journal} {J. Neurosci.}\ }\textbf {\bibinfo {volume} {26}},\
  \bibinfo {pages} {8254} (\bibinfo {year} {2006})}\BibitemShut {NoStop}%
\bibitem [{\citenamefont {Tang}\ \emph {et~al.}(2008)\citenamefont {Tang},
  \citenamefont {Jackson}, \citenamefont {Hobbs}, \citenamefont {Chen},
  \citenamefont {Smith}, \citenamefont {Patel}, \citenamefont {Prieto},
  \citenamefont {Petrusca}, \citenamefont {Grivich}, \citenamefont {Sher},
  \citenamefont {Hottowy}, \citenamefont {Dabrowski}, \citenamefont {Litke},\
  and\ \citenamefont {Beggs}}]{Tang:2008p13129}%
  \BibitemOpen
  \bibfield  {author} {\bibinfo {author} {\bibfnamefont {A.}~\bibnamefont
  {Tang}}, \bibinfo {author} {\bibfnamefont {D.}~\bibnamefont {Jackson}},
  \bibinfo {author} {\bibfnamefont {J.}~\bibnamefont {Hobbs}}, \bibinfo
  {author} {\bibfnamefont {W.}~\bibnamefont {Chen}}, \bibinfo {author}
  {\bibfnamefont {J.~L.}\ \bibnamefont {Smith}}, \bibinfo {author}
  {\bibfnamefont {H.}~\bibnamefont {Patel}}, \bibinfo {author} {\bibfnamefont
  {A.}~\bibnamefont {Prieto}}, \bibinfo {author} {\bibfnamefont
  {D.}~\bibnamefont {Petrusca}}, \bibinfo {author} {\bibfnamefont {M.~I.}\
  \bibnamefont {Grivich}}, \bibinfo {author} {\bibfnamefont {A.}~\bibnamefont
  {Sher}}, \bibinfo {author} {\bibfnamefont {P.}~\bibnamefont {Hottowy}},
  \bibinfo {author} {\bibfnamefont {W.}~\bibnamefont {Dabrowski}}, \bibinfo
  {author} {\bibfnamefont {A.~M.}\ \bibnamefont {Litke}}, \ and\ \bibinfo
  {author} {\bibfnamefont {J.~M.}\ \bibnamefont {Beggs}},\ }\href@noop {}
  {\bibfield  {journal} {\bibinfo  {journal} {Journal of Neuroscience}\ }
  (\bibinfo {year} {2008})}\BibitemShut {NoStop}%
\bibitem [{\citenamefont {Marre}\ \emph {et~al.}(2009)\citenamefont {Marre},
  \citenamefont {Boustani}, \citenamefont {Fregnac},\ and\ \citenamefont
  {Destexhe}}]{Marre:2009p13182}%
  \BibitemOpen
  \bibfield  {author} {\bibinfo {author} {\bibfnamefont {O.}~\bibnamefont
  {Marre}}, \bibinfo {author} {\bibfnamefont {S.~E.}\ \bibnamefont {Boustani}},
  \bibinfo {author} {\bibfnamefont {Y.}~\bibnamefont {Fregnac}}, \ and\
  \bibinfo {author} {\bibfnamefont {A.}~\bibnamefont {Destexhe}},\ }\href@noop
  {} {\bibfield  {journal} {\bibinfo  {journal} {Physical Review Letters}\ }
  (\bibinfo {year} {2009})}\BibitemShut {NoStop}%
\bibitem [{\citenamefont {Yu}\ \emph {et~al.}(2008)\citenamefont {Yu},
  \citenamefont {Huang}, \citenamefont {Singer},\ and\ \citenamefont
  {Nikolic}}]{Yu:2008p4351}%
  \BibitemOpen
  \bibfield  {author} {\bibinfo {author} {\bibfnamefont {S.}~\bibnamefont
  {Yu}}, \bibinfo {author} {\bibfnamefont {D.}~\bibnamefont {Huang}}, \bibinfo
  {author} {\bibfnamefont {W.}~\bibnamefont {Singer}}, \ and\ \bibinfo {author}
  {\bibfnamefont {D.}~\bibnamefont {Nikolic}},\ }\href@noop {} {\bibfield
  {journal} {\bibinfo  {journal} {Cerebral Cortex}\ } (\bibinfo {year}
  {2008})}\BibitemShut {NoStop}%
\bibitem [{\citenamefont {Broderick}\ \emph {et~al.}(2007)\citenamefont
  {Broderick}, \citenamefont {Dud\'{\i}k}, \citenamefont {Tka\v{c}ik},
  \citenamefont {Schapire},\ and\ \citenamefont
  {Bialek}}]{Broderick:2007p2761}%
  \BibitemOpen
  \bibfield  {author} {\bibinfo {author} {\bibfnamefont {T.}~\bibnamefont
  {Broderick}}, \bibinfo {author} {\bibfnamefont {M.}~\bibnamefont
  {Dud\'{\i}k}}, \bibinfo {author} {\bibfnamefont {G.}~\bibnamefont
  {Tka\v{c}ik}}, \bibinfo {author} {\bibfnamefont {R.}~\bibnamefont
  {Schapire}}, \ and\ \bibinfo {author} {\bibfnamefont {W.}~\bibnamefont
  {Bialek}},\ }\href@noop {} {\bibfield  {journal} {\bibinfo  {journal}
  {E-print arXiv}\ } (\bibinfo {year} {2007})}\BibitemShut {NoStop}%
\bibitem [{\citenamefont {MacKay}(2002)}]{mackay:book}%
  \BibitemOpen
  \bibfield  {author} {\bibinfo {author} {\bibfnamefont {D.}~\bibnamefont
  {MacKay}},\ }\href@noop {} {\emph {\bibinfo {title} {Information Theory,
  Inference and Learning Algorithms}}}\ (\bibinfo {year} {2002})\BibitemShut
  {NoStop}%
\bibitem [{\citenamefont {Chou}\ and\ \citenamefont
  {Voit}(2009)}]{genomics_review_2009}%
  \BibitemOpen
  \bibfield  {author} {\bibinfo {author} {\bibfnamefont {I.~C.}\ \bibnamefont
  {Chou}}\ and\ \bibinfo {author} {\bibfnamefont {E.~O.}\ \bibnamefont
  {Voit}},\ }\href@noop {} {\bibfield  {journal} {\bibinfo  {journal} {Math
  Biosci}\ }\textbf {\bibinfo {volume} {219}},\ \bibinfo {pages} {57} (\bibinfo
  {year} {2009})}\BibitemShut {NoStop}%
\bibitem [{\citenamefont {Aster}\ \emph {et~al.}(2005)\citenamefont {Aster},
  \citenamefont {Borchers},\ and\ \citenamefont
  {Thurber}}]{physics_param_est_book_2005}%
  \BibitemOpen
  \bibfield  {author} {\bibinfo {author} {\bibfnamefont {R.~C.}\ \bibnamefont
  {Aster}}, \bibinfo {author} {\bibfnamefont {B.}~\bibnamefont {Borchers}}, \
  and\ \bibinfo {author} {\bibfnamefont {C.~H.}\ \bibnamefont {Thurber}},\
  }\href@noop {} {\emph {\bibinfo {title} {Parameter estimation and inverse
  problems}}}\ (\bibinfo  {publisher} {Elsevier Academic Press},\ \bibinfo
  {year} {2005})\BibitemShut {NoStop}%
\bibitem [{\citenamefont {Haykin}(2008)}]{haykin2008nnc}%
  \BibitemOpen
  \bibfield  {author} {\bibinfo {author} {\bibfnamefont {S.}~\bibnamefont
  {Haykin}},\ }\href@noop {} {\emph {\bibinfo {title} {Neural networks and
  learning machines; 3rd edition}}}\ (\bibinfo  {publisher} {Prentice Hall},\
  \bibinfo {year} {2008})\BibitemShut {NoStop}%
\bibitem [{\citenamefont {Bell~AJ}(1995)}]{ICA}%
  \BibitemOpen
  \bibfield  {author} {\bibinfo {author} {\bibfnamefont {S.~T.}\ \bibnamefont
  {Bell~AJ}},\ }\href@noop {} {\bibfield  {journal} {\bibinfo  {journal}
  {Neural Computation 1995; vol. 7:1129-1159}\ } (\bibinfo {year}
  {1995})}\BibitemShut {NoStop}%
\bibitem [{\citenamefont {Cover}\ \emph {et~al.}(1991)\citenamefont {Cover},
  \citenamefont {Thomas},\ and\ \citenamefont {Wiley}}]{cover_thomas}%
  \BibitemOpen
  \bibfield  {author} {\bibinfo {author} {\bibfnamefont {T.}~\bibnamefont
  {Cover}}, \bibinfo {author} {\bibfnamefont {J.}~\bibnamefont {Thomas}}, \
  and\ \bibinfo {author} {\bibfnamefont {J.}~\bibnamefont {Wiley}},\
  }\href@noop {} {\emph {\bibinfo {title} {{Elements of information
  theory}}}},\ Vol.~\bibinfo {volume} {1}\ (\bibinfo  {publisher} {Wiley Online
  Library},\ \bibinfo {year} {1991})\BibitemShut {NoStop}%
\bibitem [{\citenamefont {Pathria}(1972)}]{Pathria:1972p5861}%
  \BibitemOpen
  \bibfield  {author} {\bibinfo {author} {\bibfnamefont {R.}~\bibnamefont
  {Pathria}},\ }\href@noop {} {\emph {\bibinfo {title} {Statistical
  Mechanics}}}\ (\bibinfo  {publisher} {Butterworth Heinemann},\ \bibinfo
  {year} {1972})\BibitemShut {NoStop}%
\bibitem [{\citenamefont {Tanaka}(1998)}]{Tanaka:1998p1984}%
  \BibitemOpen
  \bibfield  {author} {\bibinfo {author} {\bibfnamefont {T.}~\bibnamefont
  {Tanaka}},\ }\href@noop {} {\bibfield  {journal} {\bibinfo  {journal}
  {Physical Review Letters E}\ } (\bibinfo {year} {1998})}\BibitemShut
  {NoStop}%
\bibitem [{\citenamefont {Fischer~KH}(1991)}]{hertz_spin_glass}%
  \BibitemOpen
  \bibfield  {author} {\bibinfo {author} {\bibfnamefont {J.}~\bibnamefont
  {Fischer~KH}, \bibfnamefont {Hertz}},\ }\href@noop {} {\emph {\bibinfo
  {title} {Spin Glasses}}}\ (\bibinfo  {publisher} {Cambridge University
  Press},\ \bibinfo {year} {1991})\BibitemShut {NoStop}%
\bibitem [{\citenamefont {H}(2000)}]{Attias_bayes}%
  \BibitemOpen
  \bibfield  {author} {\bibinfo {author} {\bibfnamefont {A.}~\bibnamefont
  {H}},\ }\href@noop {} {\bibfield  {journal} {\bibinfo  {journal} {Advances in
  Neural Information Processing Systems 12}\ } (\bibinfo {year}
  {2000})}\BibitemShut {NoStop}%
\bibitem [{\citenamefont {Jaakkola~T}(2000)}]{Jaakkola_Jordan_bayes}%
  \BibitemOpen
  \bibfield  {author} {\bibinfo {author} {\bibfnamefont {J.~M.}\ \bibnamefont
  {Jaakkola~T}},\ }\href@noop {} {\bibfield  {journal} {\bibinfo  {journal}
  {Statistics and Computing, 10:25-37}\ } (\bibinfo {year} {2000})}\BibitemShut
  {NoStop}%
\bibitem [{\citenamefont {Besag}(1975)}]{besag}%
  \BibitemOpen
  \bibfield  {author} {\bibinfo {author} {\bibfnamefont {J.}~\bibnamefont
  {Besag}},\ }\href@noop {} {\bibfield  {journal} {\bibinfo  {journal} {The
  Statistician, 24(3), 179-195}\ } (\bibinfo {year} {1975})}\BibitemShut
  {NoStop}%
\bibitem [{\citenamefont {{Carreira-Perpi\~{n}\'an}}\ and\ \citenamefont
  {Hinton}(2004)}]{Hinton02}%
  \BibitemOpen
  \bibfield  {author} {\bibinfo {author} {\bibfnamefont {M.~A.}\ \bibnamefont
  {{Carreira-Perpi\~{n}\'an}}}\ and\ \bibinfo {author} {\bibfnamefont {G.~E.}\
  \bibnamefont {Hinton}},\ }\href@noop {} {\bibfield  {journal} {\bibinfo
  {journal} {Technical report, Dept. of Computer Science, University of
  Toronto}\ } (\bibinfo {year} {2004})}\BibitemShut {NoStop}%
\bibitem [{\citenamefont {Ackley}\ \emph {et~al.}(1985)\citenamefont {Ackley},
  \citenamefont {Hinton},\ and\ \citenamefont {Sejnowski}}]{Ackley85}%
  \BibitemOpen
  \bibfield  {author} {\bibinfo {author} {\bibfnamefont {D.~H.}\ \bibnamefont
  {Ackley}}, \bibinfo {author} {\bibfnamefont {G.~E.}\ \bibnamefont {Hinton}},
  \ and\ \bibinfo {author} {\bibfnamefont {T.~J.}\ \bibnamefont {Sejnowski}},\
  }\href@noop {} {\bibfield  {journal} {\bibinfo  {journal} {Cognitive
  Science}\ }\textbf {\bibinfo {volume} {9}},\ \bibinfo {pages} {147} (\bibinfo
  {year} {1985})}\BibitemShut {NoStop}%
\bibitem [{\citenamefont {Hyv\"arinen}(2005)}]{Hyvarinen05}%
  \BibitemOpen
  \bibfield  {author} {\bibinfo {author} {\bibfnamefont {A.}~\bibnamefont
  {Hyv\"arinen}},\ }\href@noop {} {\bibfield  {journal} {\bibinfo  {journal}
  {Journal of Machine Learning Research}\ }\textbf {\bibinfo {volume} {6}},\
  \bibinfo {pages} {695} (\bibinfo {year} {2005})}\BibitemShut {NoStop}%
\bibitem [{\citenamefont {Lyu}(2009)}]{siwei2009}%
  \BibitemOpen
  \bibfield  {author} {\bibinfo {author} {\bibfnamefont {S.}~\bibnamefont
  {Lyu}},\ }\href@noop {} {\bibfield  {journal} {\bibinfo  {journal} {The
  proceedings of the 25th conference on uncerrtainty in artificial intelligence
  (UAI*90)}\ } (\bibinfo {year} {2009})}\BibitemShut {NoStop}%
\bibitem [{\citenamefont {Movellan}(2008)}]{Movellan:2008p7643}%
  \BibitemOpen
  \bibfield  {author} {\bibinfo {author} {\bibfnamefont {J.~R.}\ \bibnamefont
  {Movellan}},\ }\href@noop {} {\bibfield  {journal} {\bibinfo  {journal}
  {unpublished draft}\ } (\bibinfo {year} {2008})}\BibitemShut {NoStop}%
\bibitem [{\citenamefont {Neal}(2010)}]{Neal:HMC}%
  \BibitemOpen
  \bibfield  {author} {\bibinfo {author} {\bibfnamefont {R.~M.}\ \bibnamefont
  {Neal}},\ }\href@noop {} {\bibfield  {journal} {\bibinfo  {journal} {Handbook
  of Markov Chain Monte Carlo}\ } (\bibinfo {year} {2010})},\ \bibinfo {note}
  {sections 5.2 and 5.3 for langevin dynamics}\BibitemShut {NoStop}%
\bibitem [{\citenamefont {Macke}\ and\ \citenamefont {Gerwinn}(2009)}]{macke}%
  \BibitemOpen
  \bibfield  {author} {\bibinfo {author} {\bibfnamefont {J.}~\bibnamefont
  {Macke}}\ and\ \bibinfo {author} {\bibfnamefont {S.}~\bibnamefont
  {Gerwinn}},\ }\href@noop {} {\bibfield  {journal} {\bibinfo  {journal}
  {Personal communication}\ } (\bibinfo {year} {2009})}\BibitemShut {NoStop}%
\bibitem [{\citenamefont {Hyv\"arinen}(2007{\natexlab{a}})}]{hyvarinen2007}%
  \BibitemOpen
  \bibfield  {author} {\bibinfo {author} {\bibfnamefont {A.}~\bibnamefont
  {Hyv\"arinen}},\ }\href@noop {} {\bibfield  {journal} {\bibinfo  {journal}
  {Computational statistics \& data analysis}\ }\textbf {\bibinfo {volume}
  {51}},\ \bibinfo {pages} {2499} (\bibinfo {year}
  {2007}{\natexlab{a}})}\BibitemShut {NoStop}%
\bibitem [{\citenamefont {Swendsen}\ and\ \citenamefont
  {Wang}(1987)}]{swendsen1987nonuniversal}%
  \BibitemOpen
  \bibfield  {author} {\bibinfo {author} {\bibfnamefont {R.}~\bibnamefont
  {Swendsen}}\ and\ \bibinfo {author} {\bibfnamefont {J.}~\bibnamefont
  {Wang}},\ }\href@noop {} {\bibfield  {journal} {\bibinfo  {journal} {Physical
  Review Letters}\ }\textbf {\bibinfo {volume} {58}},\ \bibinfo {pages} {86}
  (\bibinfo {year} {1987})}\BibitemShut {NoStop}%
\bibitem [{Note1()}]{Note1}%
  \BibitemOpen
  \bibinfo {note} {\protect \mpfisingurl}\BibitemShut {NoStop}%
\bibitem [{\citenamefont {Thouless~DJ}(1977)}]{tap_spin}%
  \BibitemOpen
  \bibfield  {author} {\bibinfo {author} {\bibfnamefont {P.~R.}\ \bibnamefont
  {Thouless~DJ}, \bibfnamefont {Anderson~PW}},\ }\href@noop {} {\bibfield
  {journal} {\bibinfo  {journal} {Philos. Mag. 35 p.593}\ } (\bibinfo {year}
  {1977})}\BibitemShut {NoStop}%
\bibitem [{\citenamefont {Dayan}\ and\ \citenamefont
  {Abbott}(2001)}]{dayan2001theoretical}%
  \BibitemOpen
  \bibfield  {author} {\bibinfo {author} {\bibfnamefont {P.}~\bibnamefont
  {Dayan}}\ and\ \bibinfo {author} {\bibfnamefont {L.}~\bibnamefont {Abbott}},\
  }\href@noop {} {\emph {\bibinfo {title} {Theoretical neuroscience}}},\
  Vol.~\bibinfo {volume} {83}\ (\bibinfo  {publisher} {Citeseer},\ \bibinfo
  {year} {2001})\BibitemShut {NoStop}%
\bibitem [{\citenamefont
  {Hyv\"arinen}(2007{\natexlab{b}})}]{Hyvarinen:2007p5984}%
  \BibitemOpen
  \bibfield  {author} {\bibinfo {author} {\bibfnamefont {A.}~\bibnamefont
  {Hyv\"arinen}},\ }\href@noop {} {\bibfield  {journal} {\bibinfo  {journal}
  {IEEE Transactions on Neural Networks}\ } (\bibinfo {year}
  {2007}{\natexlab{b}})}\BibitemShut {NoStop}%
\bibitem [{\citenamefont {Sohl-Dickstein}\ and\ \citenamefont
  {Olshausen}(2009)}]{sohldickstein}%
  \BibitemOpen
  \bibfield  {author} {\bibinfo {author} {\bibfnamefont {J.}~\bibnamefont
  {Sohl-Dickstein}}\ and\ \bibinfo {author} {\bibfnamefont {B.}~\bibnamefont
  {Olshausen}},\ }\href@noop {} {\bibfield  {journal} {\bibinfo  {journal}
  {Redwood Center Technical Report}\ } (\bibinfo {year} {2009})}\BibitemShut
  {NoStop}%
\bibitem [{\citenamefont {Welling}\ and\ \citenamefont
  {Hinton}(2002)}]{Welling:2002p3}%
  \BibitemOpen
  \bibfield  {author} {\bibinfo {author} {\bibfnamefont {M.}~\bibnamefont
  {Welling}}\ and\ \bibinfo {author} {\bibfnamefont {G.}~\bibnamefont
  {Hinton}},\ }\href@noop {} {\bibfield  {journal} {\bibinfo  {journal}
  {Lecture Notes in Computer Science}\ } (\bibinfo {year} {2002})}\BibitemShut
  {NoStop}%
\bibitem [{\citenamefont {MacKay}(2001)}]{MacKay:2001p8372}%
  \BibitemOpen
  \bibfield  {author} {\bibinfo {author} {\bibfnamefont {D.}~\bibnamefont
  {MacKay}},\ }\href@noop {} {\bibfield  {journal} {\bibinfo  {journal}
  {Failures of the one-step learning algorithm}\ } (\bibinfo {year}
  {2001})}\BibitemShut {NoStop}%
\bibitem [{\citenamefont {Yuille}(2005)}]{Yuille04}%
  \BibitemOpen
  \bibfield  {author} {\bibinfo {author} {\bibfnamefont {A.}~\bibnamefont
  {Yuille}},\ }\href@noop {} {\bibfield  {journal} {\bibinfo  {journal}
  {Department of Statistics, UCLA. Department of Statistics Papers.}\ }
  (\bibinfo {year} {2005})}\BibitemShut {NoStop}%
\bibitem [{\citenamefont {T}(1982)}]{plefka}%
  \BibitemOpen
  \bibfield  {author} {\bibinfo {author} {\bibfnamefont {P.}~\bibnamefont
  {T}},\ }\href@noop {} {\bibfield  {journal} {\bibinfo  {journal} {J. Phys. A:
  Math. Gen. 15 1971}\ } (\bibinfo {year} {1982})}\BibitemShut {NoStop}%
\bibitem [{\citenamefont {Schmidt}(2005)}]{schmidt}%
  \BibitemOpen
  \bibfield  {author} {\bibinfo {author} {\bibfnamefont {M.}~\bibnamefont
  {Schmidt}},\ }\href@noop {} {\bibfield  {journal} {\bibinfo  {journal}
  {http://www.cs.ubc.ca/~schmidtm/Software/minFunc.html}\ } (\bibinfo {year}
  {2005})}\BibitemShut {NoStop}%
\bibitem [{\citenamefont {Boyd}\ and\ \citenamefont
  {Vandenberghe}(2004)}]{boyd2004convex}%
  \BibitemOpen
  \bibfield  {author} {\bibinfo {author} {\bibfnamefont {S.}~\bibnamefont
  {Boyd}}\ and\ \bibinfo {author} {\bibfnamefont {L.}~\bibnamefont
  {Vandenberghe}},\ }\href@noop {} {\emph {\bibinfo {title} {{Convex
  optimization}}}}\ (\bibinfo  {publisher} {Cambridge Univ Pr},\ \bibinfo
  {year} {2004})\BibitemShut {NoStop}%
\bibitem [{\citenamefont {Brush}(1967)}]{RevModPhys.39.883}%
  \BibitemOpen
  \bibfield  {author} {\bibinfo {author} {\bibfnamefont {S.~G.}\ \bibnamefont
  {Brush}},\ }\href@noop {} {\bibfield  {journal} {\bibinfo  {journal} {Reviews
  of Modern Physics}\ }\textbf {\bibinfo {volume} {39}},\ \bibinfo {pages}
  {883} (\bibinfo {year} {1967})}\BibitemShut {NoStop}%
\bibitem [{\citenamefont {Hateren}\ and\ \citenamefont
  {Schaaf}(1998)}]{hateren_schaaf_1998}%
  \BibitemOpen
  \bibfield  {author} {\bibinfo {author} {\bibfnamefont {J.~H.~v.}\
  \bibnamefont {Hateren}}\ and\ \bibinfo {author} {\bibfnamefont {A.~v.~d.}\
  \bibnamefont {Schaaf}},\ }\href@noop {} {\bibfield  {journal} {\bibinfo
  {journal} {Proceedings: Biological Sciences}\ }\textbf {\bibinfo {volume}
  {265}},\ \bibinfo {pages} {359} (\bibinfo {year} {1998})}\BibitemShut
  {NoStop}%
\end{thebibliography}%


%Merlin.mbs v4.21 2009-07-09.
\begin{thebibliography}{10}%
\makeatletter
\providecommand \@ifxundefined [1]{%
 \ifx #1\undefined \expandafter \@firstoftwo
 \else \expandafter \@secondoftwo
\fi
}%
\providecommand \@ifnum [1]{%
 \ifnum #1\expandafter \@firstoftwo
 \else \expandafter \@secondoftwo
\fi
}%
\providecommand \enquote [1]{``#1''}%
\providecommand \bibnamefont  [1]{#1}%
\providecommand \bibfnamefont [1]{#1}%
\providecommand \citenamefont [1]{#1}%
\providecommand\href[0]{\@sanitize\@href}%
\providecommand\@href[1]{\endgroup\@@startlink{#1}\endgroup\@@href}%
\providecommand\@@href[1]{#1\@@endlink}%
\providecommand \@sanitize [0]{\begingroup\catcode`\&12\catcode`\#12\relax}%
\@ifxundefined \pdfoutput {\@firstoftwo}{%
 \@ifnum{\z@=\pdfoutput}{\@firstoftwo}{\@secondoftwo}%
}{%
 \providecommand\@@startlink[1]{\leavevmode\special{html:<a href="#1">}}%
 \providecommand\@@endlink[0]{\special{html:</a>}}%
}{%
 \providecommand\@@startlink[1]{%
  \leavevmode
  \pdfstartlink
   attr{/Border[0 0 1 ]/H/I/C[0 1 1]}%
   user{/Subtype/Link/A<</Type/Action/S/URI/URI(#1)>>}%
  \relax
 }%
 \providecommand\@@endlink[0]{\pdfendlink}%
}%
\providecommand \url  [0]{\begingroup\@sanitize \@url }%
\providecommand \@url [1]{\endgroup\@href {#1}{\urlprefix}}%
\providecommand \urlprefix [0]{URL }%
\providecommand \Eprint[0]{\href }%
\@ifxundefined \urlstyle {%
  \providecommand \doi [1]{doi:\discretionary{}{}{}#1}%
}{%
  \providecommand \doi [0]{doi:\discretionary{}{}{}\begingroup
  \urlstyle{rm}\Url }%
}%
\providecommand \doibase [0]{http://dx.doi.org/}%
\providecommand \Doi[1]{\href{\doibase#1}}%
\providecommand \bibAnnote [3]{%
  \BibitemShut{#1}%
  \begin{quotation}\noindent
    \textsc{Key:}\ #2\\\textsc{Annotation:}\ #3%
  \end{quotation}%
}%
\providecommand \bibAnnoteFile [2]{%
  \IfFileExists{#2}{\bibAnnote {#1} {#2} {\input{#2}}}{}%
}%
\providecommand \typeout [0]{\immediate \write \m@ne }%
\providecommand \selectlanguage [0]{\@gobble}%
\providecommand \bibinfo [0]{\@secondoftwo}%
\providecommand \bibfield [0]{\@secondoftwo}%
\providecommand \translation [1]{[#1]}%
\providecommand \BibitemOpen[0]{}%
\providecommand \bibitemStop [0]{}%
\providecommand \bibitemNoStop [0]{.\EOS\space}%
\providecommand \EOS [0]{\spacefactor3000\relax}%
\providecommand \BibitemShut [1]{\csname bibitem#1\endcsname}%
%</preamble>
\bibitem{macke}%
  \BibitemOpen
  \bibfield{author}{%
  \bibinfo {author} {\bibfnamefont{J.}~\bibnamefont{Macke}}\ and\ \bibinfo
  {author} {\bibfnamefont{S.}~\bibnamefont{Gerwinn}},\ }%
  \bibfield{journal}{%
  \bibinfo {journal} {Personal communication}}%
   (\bibinfo {year} {2009})%
  \bibAnnoteFile{NoStop}{macke}%
\bibitem{Hyvarinen05}%
  \BibitemOpen
  \bibfield{author}{%
  \bibinfo {author} {\bibfnamefont{A.}~\bibnamefont{Hyv\"arinen}},\ }%
  \bibfield{journal}{%
  \bibinfo {journal} {Journal of Machine Learning Research}\ }%
  \textbf{\bibinfo {volume} {6}},\ \bibinfo {pages} {695} (\bibinfo {year}
  {2005})%
  \bibAnnoteFile{NoStop}{Hyvarinen05}%
\bibitem{Hyvarinen:2007p5984}%
  \BibitemOpen
  \bibfield{author}{%
  \bibinfo {author} {\bibfnamefont{A.}~\bibnamefont{Hyv\"arinen}},\ }%
  \bibfield{journal}{%
  \bibinfo {journal} {IEEE Transactions on Neural Networks}}%
   (\bibinfo {month} {Jan}\ \bibinfo {year} {2007})%
  \bibAnnoteFile{NoStop}{Hyvarinen:2007p5984}%
\bibitem{sohldickstein}%
  \BibitemOpen
  \bibfield{author}{%
  \bibinfo {author} {\bibfnamefont{J.}~\bibnamefont{Sohl-Dickstein}}\ and\
  \bibinfo {author} {\bibfnamefont{B.}~\bibnamefont{Olshausen}},\ }%
  \bibfield{journal}{%
  \bibinfo {journal} {Redwood Center Technical Report}}%
   (\bibinfo {year} {2009})%
  \bibAnnoteFile{NoStop}{sohldickstein}%
\bibitem{Movellan:2008p7643}%
  \BibitemOpen
  \bibfield{author}{%
  \bibinfo {author} {\bibfnamefont{J.~R.}\ \bibnamefont{Movellan}},\ }%
  \bibfield{journal}{%
  \bibinfo {journal} {unpublished draft}}%
   (\bibinfo {month} {Jan}\ \bibinfo {year} {2008})%
  \bibAnnoteFile{NoStop}{Movellan:2008p7643}%
\bibitem{siwei2009}%
  \BibitemOpen
  \bibfield{author}{%
  \bibinfo {author} {\bibfnamefont{S.}~\bibnamefont{Lyu}},\ }%
  \bibfield{journal}{%
  \bibinfo {journal} {The proceedings of the 25th conference on uncerrtainty in
  artificial intelligence (UAI*90)}}%
   (\bibinfo {year} {2009})%
  \bibAnnoteFile{NoStop}{siwei2009}%
\bibitem{hyvarinen2007}%
  \BibitemOpen
  \bibfield{author}{%
  \bibinfo {author} {\bibfnamefont{A.}~\bibnamefont{Hyv\"arinen}},\ }%
  \bibfield{journal}{%
  \bibinfo {journal} {Computational statistics \& data analysis}\ }%
  \textbf{\bibinfo {volume} {51}},\ \bibinfo {pages} {2499} (\bibinfo {year}
  {2007}),\ ISSN \bibinfo {issn} {0167-9473}%
  \bibAnnoteFile{NoStop}{hyvarinen2007}%
\bibitem{Welling:2002p3}%
  \BibitemOpen
  \bibfield{author}{%
  \bibinfo {author} {\bibfnamefont{M.}~\bibnamefont{Welling}}\ and\ \bibinfo
  {author} {\bibfnamefont{G.}~\bibnamefont{Hinton}},\ }%
  \bibfield{journal}{%
  \bibinfo {journal} {Lecture Notes in Computer Science}}%
   (\bibinfo {month} {Jan}\ \bibinfo {year} {2002})%
  \bibAnnoteFile{NoStop}{Welling:2002p3}%
\bibitem{Hinton02}%
  \BibitemOpen
  \bibfield{author}{%
  \bibinfo {author} {\bibfnamefont{M.~A.}\
  \bibnamefont{{Carreira-Perpi\~{n}\'an}}}\ and\ \bibinfo {author}
  {\bibfnamefont{G.~E.}\ \bibnamefont{Hinton}},\ }%
  \bibfield{journal}{%
  \bibinfo {journal} {Technical report, Dept. of Computer Science, University
  of Toronto}}%
   (\bibinfo {year} {2004})%
  \bibAnnoteFile{NoStop}{Hinton02}%
\bibitem{MacKay:2001p8372}%
  \BibitemOpen
  \bibfield{author}{%
  \bibinfo {author} {\bibfnamefont{D.}~\bibnamefont{MacKay}},\ }%
  \bibfield{journal}{%
  \bibinfo {journal} {Failures of the one-step learning algorithm}}%
   (\bibinfo {month} {Jan}\ \bibinfo {year} {2001})%
  \bibAnnoteFile{NoStop}{MacKay:2001p8372}%
\bibitem{Yuille04}%
  \BibitemOpen
  \bibfield{author}{%
  \bibinfo {author} {\bibfnamefont{A.}~\bibnamefont{Yuille}},\ }%
  \bibfield{journal}{%
  \bibinfo {journal} {Department of Statistics, UCLA. Department of Statistics
  Papers.}}%
   (\bibinfo {year} {2005})%
  \bibAnnoteFile{NoStop}{Yuille04}%
\bibitem{tap_spin}%
  \BibitemOpen
  \bibfield{author}{%
  \bibinfo {author} {\bibfnamefont{P.~R.}\ \bibnamefont{Thouless~DJ},
  \bibfnamefont{Anderson~PW}},\ }%
  \bibfield{journal}{%
  \bibinfo {journal} {Philos. Mag. 35 p.593}}%
   (\bibinfo {year} {1977})%
  \bibAnnoteFile{NoStop}{tap_spin}%
\bibitem{besag}%
  \BibitemOpen
  \bibfield{author}{%
  \bibinfo {author} {\bibfnamefont{J.}~\bibnamefont{Besag}},\ }%
  \bibfield{journal}{%
  \bibinfo {journal} {The Statistician, 24(3), 179-195}}%
   (\bibinfo {year} {1975})%
  \bibAnnoteFile{NoStop}{besag}%
\bibitem{plefka}%
  \BibitemOpen
  \bibfield{author}{%
  \bibinfo {author} {\bibfnamefont{P.}~\bibnamefont{T}},\ }%
  \bibfield{journal}{%
  \bibinfo {journal} {J. Phys. A: Math. Gen. 15 1971}}%
   (\bibinfo {year} {1982})%
  \bibAnnoteFile{NoStop}{plefka}%
\bibitem{hertz_spin_glass}%
  \BibitemOpen
  \bibfield{author}{%
  \bibinfo {author} {\bibfnamefont{J.}~\bibnamefont{Fischer~KH},
  \bibfnamefont{Hertz}},\ }%
  \emph{\bibinfo {title} {Spin Glasses}}\ (\bibinfo {publisher} {Cambridge
  University Press},\ \bibinfo {year} {1991})%
  \bibAnnoteFile{NoStop}{hertz_spin_glass}%
\bibitem{schmidt}%
  \BibitemOpen
  \bibfield{author}{%
  \bibinfo {author} {\bibfnamefont{M.}~\bibnamefont{Schmidt}},\ }%
\\  \bibfield{journal}{
  \bibinfo {journal} {{\footnotesize http://www.cs.ubc.ca/$\sim$schmidtm/Software/minFunc.html}}} 
   (\bibinfo {year} {2005})%
  \bibAnnoteFile{NoStop}{schmidt}%
\bibitem{boyd2004convex}%
  \BibitemOpen
  \bibfield{author}{%
  \bibinfo {author} {\bibfnamefont{S.}~\bibnamefont{Boyd}}\ and\ \bibinfo
  {author} {\bibfnamefont{L.}~\bibnamefont{Vandenberghe}},\ }%
  \emph{\bibinfo {title} {{Convex optimization}}}\ (\bibinfo {publisher}
  {Cambridge Univ Pr},\ \bibinfo {year} {2004})\ ISBN \bibinfo {isbn}
  {0521833787}%
  \bibAnnoteFile{NoStop}{boyd2004convex}%
\bibitem{RevModPhys.39.883}%
  \BibitemOpen
  \bibfield{author}{%
  \bibinfo {author} {\bibfnamefont{S.~G.}\ \bibnamefont{Brush}},\ }%
  \bibfield{journal}{%
  \bibinfo {journal} {Reviews of Modern Physics}\ }%
  \textbf{\bibinfo {volume} {39}},\ \bibinfo {pages} {883} (\bibinfo {month}
  {Oct}\ \bibinfo {year} {1967})%
  \bibAnnoteFile{NoStop}{RevModPhys.39.883}%
\bibitem{Ackley85}%
  \BibitemOpen
  \bibfield{author}{%
  \bibinfo {author} {\bibfnamefont{D.~H.}\ \bibnamefont{Ackley}}, \bibinfo
  {author} {\bibfnamefont{G.~E.}\ \bibnamefont{Hinton}},\ and\ \bibinfo
  {author} {\bibfnamefont{T.~J.}\ \bibnamefont{Sejnowski}},\ }%
  \bibfield{journal}{%
  \bibinfo {journal} {Cognitive Science}\ }%
  \textbf{\bibinfo {volume} {9}},\ \bibinfo {pages} {147} (\bibinfo {year}
  {1985})%
  \bibAnnoteFile{NoStop}{Ackley85}%
\bibitem{Schneidman_Nature_2006}%
  \BibitemOpen
  \bibfield{author}{%
  \bibinfo {author} {\bibfnamefont{E.}~\bibnamefont{Schneidman}}, \bibinfo
  {author} {\bibfnamefont{M.~J.~B.}\ \bibnamefont{2nd}}, \bibinfo {author}
  {\bibfnamefont{R.}~\bibnamefont{Segev}},\ and\ \bibinfo {author}
  {\bibfnamefont{W.}~\bibnamefont{Bialek}},\ }%
  \bibfield{journal}{%
  \bibinfo {journal} {Nature}\ }%
  \textbf{\bibinfo {volume} {440}},\ \bibinfo {pages} {1007} (\bibinfo {year}
  {2006})%
  \bibAnnoteFile{NoStop}{Schneidman_Nature_2006}%
\bibitem{Shlens_JN_2006}%
  \BibitemOpen
  \bibfield{author}{%
  \bibinfo {author} {\bibfnamefont{J.}~\bibnamefont{Shlens}}, \bibinfo {author}
  {\bibfnamefont{G.~D.}\ \bibnamefont{Field}}, \bibinfo {author}
  {\bibfnamefont{J.~L.}\ \bibnamefont{Gauthier}}, \bibinfo {author}
  {\bibfnamefont{M.~I.}\ \bibnamefont{Grivich}}, \bibinfo {author}
  {\bibfnamefont{D.}~\bibnamefont{Petrusca}}, \bibinfo {author}
  {\bibfnamefont{A.}~\bibnamefont{Sher}}, \bibinfo {author}
  {\bibfnamefont{A.~M.}\ \bibnamefont{Litke}},\ and\ \bibinfo {author}
  {\bibfnamefont{E.~J.}\ \bibnamefont{Chichilnisky}},\ }%
  \bibfield{journal}{%
  \bibinfo {journal} {J. Neurosci.}\ }%
  \textbf{\bibinfo {volume} {26}},\ \bibinfo {pages} {8254} (\bibinfo {year}
  {2006})%
  \bibAnnoteFile{NoStop}{Shlens_JN_2006}%
\bibitem{Broderick:2007p2761}%
  \BibitemOpen
  \bibfield{author}{%
  \bibinfo {author} {\bibfnamefont{T.}~\bibnamefont{Broderick}}, \bibinfo
  {author} {\bibfnamefont{M.}~\bibnamefont{Dud\'{\i}k}}, \bibinfo {author}
  {\bibfnamefont{G.}~\bibnamefont{Tka\v{c}ik}}, \bibinfo {author}
  {\bibfnamefont{R.}~\bibnamefont{Schapire}},\ and\ \bibinfo {author}
  {\bibfnamefont{W.}~\bibnamefont{Bialek}},\ }%
  \bibfield{journal}{%
  \bibinfo {journal} {E-print arXiv}}%
   (\bibinfo {month} {Jan}\ \bibinfo {year} {2007})%
  \bibAnnoteFile{NoStop}{Broderick:2007p2761}%
\bibitem{ICA}%
  \BibitemOpen
  \bibfield{author}{%
  \bibinfo {author} {\bibfnamefont{S.~T.}\ \bibnamefont{Bell~AJ}},\ }%
  \bibfield{journal}{%
  \bibinfo {journal} {Neural Computation 1995; vol. 7:1129-1159}}%
   (\bibinfo {year} {1995})%
  \bibAnnoteFile{NoStop}{ICA}%
\bibitem{hateren_schaaf_1998}%
  \BibitemOpen
  \bibfield{author}{%
  \bibinfo {author} {\bibfnamefont{J.~H.~v.}\ \bibnamefont{Hateren}}\ and\
  \bibinfo {author} {\bibfnamefont{A.~v.~d.}\ \bibnamefont{Schaaf}},\ }%
  \bibfield{journal}{%
  \bibinfo {journal} {Proceedings: Biological Sciences}\ }%
  \textbf{\bibinfo {volume} {265}},\ \bibinfo {pages} {359} (\bibinfo {month}
  {Mar}\ \bibinfo {year} {1998})%
  \bibAnnoteFile{NoStop}{hateren_schaaf_1998}%
\bibitem{Neal:HMC}%
  \BibitemOpen
  \bibfield{author}{%
  \bibinfo {author} {\bibfnamefont{R.~M.}\ \bibnamefont{Neal}},\ }%
  \bibfield{journal}{%
  \bibinfo {journal} {Handbook of Markov Chain Monte Carlo}}%
   (\bibinfo {month} {Jan}\ \bibinfo {year} {2010}),\ \bibinfo {note} {sections
  5.2 and 5.3 for langevin dynamics}%
  \bibAnnoteFile{NoStop}{Neal:HMC}%
\end{thebibliography}%


%merlin.mbs apsrev4-1.bst 2010-07-25 4.21a (PWD, AO, DPC) hacked
%Control: key (0)
%Control: author (8) initials jnrlst
%Control: editor formatted (1) identically to author
%Control: production of article title (-1) disabled
%Control: page (0) single
%Control: year (1) truncated
%Control: production of eprint (0) enabled
%

\newpage
~
\newpage

%%\documentclass[10pt]{article}
%\documentclass[aps,showpacs,twocolumn,groupedaddress]{revtex4-1}  % for review and submission
%
%\newcommand{\set}[1]{\lbrace #1 \rbrace}
%\newcommand{\setc}[2]{\lbrace #1 \mid #2 \rbrace}
%\newcommand{\vv}[1]{{\mb{#1}}}
%\newcommand{\dd}{{\mathrm{d}}}
%\newcommand{\pd}[2]{\frac{\partial #1}{\partial #2}}
%\newcommand{\pdn}[3]{\frac{\partial^#1 #2}{\partial #3^#1}}
%\newcommand{\od}[2]{\frac{\dd #1}{\dd #2}}
%\newcommand{\odn}[3]{\frac{\dd^#1 #2}{\dd #3^#1}}
%\newcommand{\avg}[1]{\left< #1 \right>}
%\newcommand{\p}[2]{p_{#1}^{(#2)}}
%\newcommand{\mb}{\mathbf}
%\newcommand{\argmin}{\operatornamewithlimits{argmin}}
%\newcommand{\doublebar}{\bigl|\!\bigr|}
%
%
%
%\newcommand{\ham}[2]{\operatornamewithlimits{HAM}\left( #1 ; #2 \right)}
%
%% For figures
%\usepackage{amsmath,amssymb}
%\usepackage{graphicx}
%
%
%\begin{document}

%\begin{center}
%	{\bf APPENDICES}
 %\end{center}

\appendix

%\section{Supplemental Material}
%\maketitle

\section{Taylor Expansion of KL Divergence}
\label{app:KL}

\renewcommand{\theequation}{A-\arabic{equation}}
 % redefine the command that creates the equation no.
 \setcounter{equation}{0}  % reset counter

The minimum probability flow learning objective function $K\left(\theta\right)$ is found by taking up to the first order terms in the Taylor expansion of the KL divergence between the data distribution and the distribution resulting from running the dynamics for a time $\epsilon$: 
\begin{align}
K\left( \theta \right) & \approx  D_{KL}\left(
   \mb{p^{(0)}} ||\mb{p^{(t)}}
   \left(\theta\right)\right)\Big |_{t=0}
\nonumber \\ & \qquad
   + \epsilon \frac
   {\partial D_{KL}\left(
   \mb{p^{(0)}} ||\mb{p^{(t)}}
   \left(\theta\right)\right)}
   {\partial t}\Big |_{t=0} \\
&= 0
   + \epsilon \frac
   {\partial D_{KL}\left(
   \mb{p^{(0)}} ||\mb{p^{(t)}}
   \left(\theta\right)\right)}
   {\partial t}\Big |_{t=0} \\
&= \epsilon \pd{}{t}\left.\left(\sum_{i\in \mathcal{D}} \p{i}{0}\log\frac{\p{i}{0}}{\p{i}{t}}\right)\right|_0 \\
&= -\epsilon \sum_{i\in \mathcal{D}}\frac{\p{i}{0}}{\p{i}{0}}\left.\pd{\p{i}{t}}{t}\right|_0 \\
&= -\epsilon \left.\sum_{i\in \mathcal{D}}\pd{\p{i}{t}}{t}\right|_0 \\\displaybreak[0]
&= -\epsilon \left.\left(\pd{}{t}\sum_{i\in \mathcal{D}}\p{i}{t}\right)\right|_0  \label{eqn:sumrate} \\ \displaybreak[0]
&= -\epsilon \left.\pd{}{t}\left(1-\sum_{i\notin \mathcal{D}}\p{i}{t}\right)\right|_0 \\ \displaybreak[0]
&= \epsilon \left.\sum_{i\notin \mathcal{D}}\pd{\p{i}{t}}{t}\right|_0 \\\displaybreak[0]
&= \epsilon \sum_{i\notin \mathcal{D}}\sum_{j\in \mathcal{D}}\Gamma_{ij}\p{j}{0} \\
&= \frac{\epsilon}{|\mathcal{D}|} \sum_{i\notin \mathcal{D}}\sum_{j\in \mathcal{D}}\Gamma_{ij},
\end{align}
where we used the fact that $\sum_{i\in \mathcal{D}} p_i^{(t)} + \sum_{i\notin \mathcal{D}} p_i^{(t)} = 1$.
This implies that the rate of growth of the KL divergence at time $t=0$ equals the total initial flow of probability from states with data into states without.

\section{Convexity}
\label{app:convex}

\renewcommand{\theequation}{B-\arabic{equation}}
 % redefine the command that creates the equation no.
 \setcounter{equation}{0}  % reset counter

As observed by Macke and Gerwinn \cite{macke}, the MPF objective function is convex for models in the exponential family.

We wish to minimize
\begin{align}
K &=
   \sum_{i \in D} \sum_{j \in D^C} \Gamma_{ji} p_i^{(0)} .
\end{align}

$K$ has derivative
\begin{align}
\pd{K}{\theta_m} &= \sum_{i \in D}\sum_{j \in D^c} \left( \pd{\Gamma_{ij}}{\theta_m} \right) p_i^{(0)} \\
&= \frac{1}{2} \sum_{i \in D}\sum_{j \in D^c} \Gamma_{ij} \left(  \pd{E_j}{\theta_m} - \pd{E_i}{\theta_m}  \right) p_i^{(0)},
\end{align}
and Hessian
\begin{align}
\pd{^2 K}{\theta_m \partial \theta_n}
   &=
   	\frac{1}{4} \sum_{i \in D}\sum_{j \in D^c} \Gamma_{ij} \left(  \pd{E_j}{\theta_m} - \pd{E_i}{\theta_m}  \right)\left(  \pd{E_j}{\theta_n} - \pd{E_i}{\theta_n}  \right) p_i^{(0)}  \nonumber \\
   	&+
   	\frac{1}{2} \sum_{i \in D}\sum_{j \in D^c} \Gamma_{ij} \left(  \pd{^2 E_j}{\theta_m \partial \theta_n} - \pd{^2 E_i}{\theta_m \partial \theta_n}  \right) p_i^{(0)} .
\end{align}
The first term in the Hessian is a weighted sum of outer products, with non-negative weights $\frac{1}{4} \Gamma_{ij} p_i^{(0)}$, and is thus positive semidefinite.  The second term is $0$ for models in the exponential family (those with energy functions linear in their parameters).

Parameter estimation for models in the exponential family is therefore convex using minimum probability flow learning.

\section{Relationship of MPF to other techniques}
\subsection{Score matching}
\label{app:score_matching}

\renewcommand{\theequation}{C-\arabic{equation}}
% redefine the command that creates the equation no.
\setcounter{equation}{0}  % reset counter

Score matching, developed by Aapo Hyv\"arinen \cite{Hyvarinen05}, is a method that learns parameters in a probabilistic model using only derivatives of the energy function evaluated over the data distribution (see Equation \eqref{eq:score matching}).  This sidesteps the need to explicitly sample or integrate over the model distribution. In score matching one minimizes the expected square distance of the score function with respect to spatial coordinates given by the data distribution from the similar score function given by the model distribution.  A number of connections have been made between score matching and other learning techniques \cite{Hyvarinen:2007p5984, sohldickstein, Movellan:2008p7643, siwei2009}.  Here we show that in the correct limit, MPF also reduces to score matching.

For a $d$-dimensional, continuous state space, we can write the MPF objective function as
\begin{align}
K_{\mathrm{MPF}} & = \frac{1}{N}\sum_{x\in\mathcal{D}} \int \dd^d y\; \Gamma(y,x)\nonumber \\
&= \frac{1}{N}\sum_{x\in\mathcal{D}} \int \dd^d y\; g(y,x) e^{\frac{1}{2}(E(x|\theta)-E(y|\theta))},
\end{align}
where the sum $\sum_{x\in\mathcal{D}}$ is over all data samples, and $N$ is the number of samples in the data set $\mathcal{D}$.
Now we assume that transitions are only allowed from states $x$ to states $y$ that are within
a hypercube of side length $\epsilon$ centered around $x$ in state space. (The master equation
will reduce to Gaussian diffusion as $\epsilon\rightarrow 0$.) Thus, the function $g(y,x)$ will
equal 1 when $y$ is within the $x$-centered cube (or $x$ within the $y$-centered cube) and 0 otherwise. Calling this cube $C_\epsilon$,
and writing $y=x+\alpha$ with $\alpha\in C_\epsilon$, we have
\begin{align}
K_{\mathrm{MPF}} = \frac{1}{N}\sum_{x\in\mathcal{D}} \int_{C_\epsilon} \dd^d \alpha\;
e^{\frac{1}{2}(E(x|\theta)-E(x+\alpha|\theta))}.
\end{align}
If we Taylor expand in $\alpha$ to second order and ignore cubic and higher terms, we get
\begin{align}
K_{\mathrm{MPF}} &\approx \frac{1}{N}\sum_{x\in\mathcal{D}} \int_{C_\epsilon} \dd^d\alpha\;
(1) \nonumber\\
&- \frac{1}{N}\sum_{x\in\mathcal{D}} \int_{C_\epsilon} \dd^d\alpha\;
\frac{1}{2}\sum_{i=1}^d\alpha_i\nabla_{x_i}E(x|\theta)\nonumber \\
&+\frac{1}{N}\sum_{x\in\mathcal{D}} \int_{C_\epsilon} \dd^d\alpha\;
\frac{1}{4}\Biggl(\frac{1}{2}\biggl[\sum_{i=1}^d
\alpha_i\nabla_{x_i}E(x|\theta)\biggr]^2 \nonumber \\
&\quad-\sum_{i,j=1}^d\alpha_i\alpha_j\nabla_{x_i}\nabla_{x_j}E(x|\theta)\Biggr).
\end{align}
This reduces to
\begin{align}
K_{\mathrm{MPF}}&\approx\frac{1}{N}\sum_{x\in\mathcal{D}}\Biggl[\epsilon^d + \frac{1}{4}\Biggl(\frac{1}{2}\frac{1}{12}\epsilon^{d+2}
\sum_{i=1}^d\biggl[\nabla_{x_i}E(x|\theta)\biggr]^2 \nonumber \\ &\quad-\frac{1}{12}\epsilon^{d+2}\sum_{i=1}^d\nabla_{x_i}^2E(x|\theta)\Biggr)\Biggr],
\end{align}
which, removing a constant offset and scaling factor, is exactly equal to the score matching objective function,
\begin{align}
K_{\mathrm{MPF}} &\sim \frac{1}{N}\sum_{x\in\mathcal{D}}
\biggl[ \frac{1}{2}\nabla E(x|\theta)\cdot\nabla E(x|\theta) -\nabla^2E(x|\theta)\biggr]\label{eq:score matching}\\
&= K_{\mathrm{SM}}
.
\end{align}
Score matching is thus equivalent to MPF when the connectivity function $g(y,x)$ is non-zero only for states infinitesimally close to each other. It should be noted that the score matching estimator has a closed-form solution when the model distribution belongs to the exponential family \cite{hyvarinen2007}, so the same can be said for MPF in this limit.

\subsection{Contrastive divergence}
\label{app:contrastive_divergence}

Contrastive divergence \cite{Welling:2002p3,Hinton02} is a variation on steepest gradient descent of the maximum (log) likelihood (ML) objective function.  Rather than integrating over the full model distribution, CD approximates the partition function term in the gradient by averaging over the distribution obtained after taking a few, or only one, Markov chain Monte Carlo (MCMC) step away from the data distribution (Equation \ref{eq:CDstep}).
Qualitatively, one can imagine that the data distribution is contrasted against a distribution which has evolved only a small distance towards the model distribution, whereas it would be contrasted against the true model distribution in traditional MCMC approaches.  Although CD is not guaranteed to converge to the right answer, or even to a fixed point, it has proven to be an effective and fast heuristic for parameter estimation \cite{MacKay:2001p8372,Yuille04}.

The contrastive divergence update rule can be written in the form
\begin{align}
\label{eq:CDstep}
\Delta \theta_{CD} & \propto
 -\sum_{j\in \mathrm{\mathcal{D}}
} \sum_{
      i\notin \mathrm{\mathcal{D}}
}
\left[ \pd{E_j\left( \theta \right)}{\theta}-\pd{E_i\left(  \theta \right)}{\theta} \right]
T_{ij}
,
\end{align}
where $T_{ij}$ is the probability of transitioning from state $j$ to state $i$ in a single Markov chain Monte Carlo step (or a small number of steps).  Equation \ref{eq:CDstep} has obvious similarities to the MPF learning gradient
\begin{eqnarray}
\label{eq:Kgrad}
\pd{K\left( \theta \right) }{\theta}
&= \frac{\epsilon}{2N} \sum_{j\in \mathrm{\mathcal{D}}} \sum_{i\notin \mathrm{\mathcal{D}}}
\left[ \pd{E_j\left( \theta \right)}{\theta}-\pd{E_i\left(  \theta \right)}{\theta} \right] \\ 
&\qquad g_{ij} \exp \left[ \frac{1}{2} \left( E_j\left( \theta \right)-E_i\left(  \theta \right) \right)\right] 
.
\end{eqnarray}
Thus, steepest gradient descent under MPF resembles CD updates, but with the MCMC sampling/rejection step $T_{ij}$ replaced by a weighting factor $g_{ij} \exp \left[ \frac{1}{2} \left( E_j\left( \theta \right)-E_i\left(  \theta \right) \right)\right]$.  Note that this difference in form provides MPF with a well-defined objective function, and it guarantees consistency ({\em i.e.}, there is a global minimum when model and data distributions agree).

\renewcommand{\theequation}{D-\arabic{equation}}
% redefine the command that creates the equation no.
\setcounter{equation}{0}  % reset counter

\section{Sampling the connectivity matrix $\mathbf \Gamma$}
\renewcommand{\theequation}{D-\arabic{equation}}
 % redefine the command that creates the equation no.
 \setcounter{equation}{0}  % reset counter
\renewcommand{\thefigure}{D-\arabic{figure}}
 \setcounter{figure}{0}  % reset fig counter

The MPF learning scheme is  blind to regions in state space which data states don't have any connectivity to - the flow at time 0 is only a function of the states that are directly connected to data states.  To get the most informative learning signal, it seems sensible to encourage probability flow directly between data states and states that are probable under the model.  That way the objective function is sensitive to the regions which are probable under the model.  We believe nearest neighbor connectivity schemes are effective largely because as the parameters converge the regions around data states become the high probability regions for the model.  We wish to try connectivity schemes other than nearest neighbors to allow probability to most efficiently flow between data states and high probability model states.  In order to do so we need to slightly extend the MPF algorithm.  We do this by allowing the connectivity pattern in $\Gamma$ to be sampled independently in every infinitesimal time step.

Since $\Gamma$ is now sampled, we will modify detailed balance to demand that, averaging over the choices for $\Gamma$, the net flow between pairs of states is 0.
\begin{eqnarray}
\label{eq:average detailed balance}
\left< \Gamma_{ji} \  p^{(\infty)}_i\left(\theta\right) \right> & = & \left< \Gamma_{ij} \  p^{(\infty)}_j\left(\theta\right) \right> \\
\left< \Gamma_{ji} \right>\  p^{(\infty)}_i\left(\theta\right)  & = & \left< \Gamma_{ij} \right>\  p^{(\infty)}_j\left(\theta\right) 
,
\end{eqnarray}
where the ensemble average is over the connectivity scheme for $\Gamma$.
  We describe the connectivity scheme via a function $g_{ij}$, such that the probability of there being a connection from state $j$ to state $i$ at any given moment is $g_{ij}$.  We also introduce a function $F_{ij}$, which provides the value $\Gamma_{ij}$ takes on when a connection occurs from $j$ to $i$.  That is, it is the probability flow rate when flow occurs -
\begin{equation}
\label{eq:detailed_balance}
\left< \Gamma_{ij} \right> = g_{ij} F_{ij}
.
\end{equation}
Detailed balance now becomes
\begin{equation}
\label{eq:detailed_balance}
g_{ji} F_{ji} \  p^{(\infty)}_i\left(\theta\right) = g_{ij} F_{ij} \  p^{(\infty)}_j\left(\theta\right)
.
\end{equation}
Solving for $\mathbf F$ we find
\begin{equation}
\frac
	{F_{ij}}
	{F_{ji}}
=
\frac
	{g_{ji}}
	{g_{ij}}
\frac
	{p^{(\infty)}_i\left(\theta\right)}
	{p^{(\infty)}_j\left(\theta\right)}
=
\frac
	{g_{ji}}
	{g_{ij}}
	\exp \left[ E_j\left( \theta \right)-E_i\left( \theta \right) \right]
.
\end{equation}
$\mathbf{F}$ is underconstrained by the above equation. Motivated by symmetry and aesthetics, we choose as the form for the (non-zero, non-diagonal) entries in $\mathbf{F}$
\begin{eqnarray}
\label{eqn:F symmetric}
F_{ij} = 
\left( \frac
	{g_{ji}}
	{g_{ij}}
\right)^{\frac{1}{2}}
	\exp \left[ \frac{1}{2} \left( E_j\left( \theta \right)-E_i\left(  \theta \right) \right)\right]
.
\end{eqnarray}

$\mathbf\Gamma$ is now populated as
\begin{eqnarray}
r_{ij} & \sim & \mathrm{ rand\ [0, 1) }
\\
\Gamma_{ij} & = &
	\left\{\begin{array}{ccc}
		-\sum_{k \neq i}\Gamma_{ki} &  & i = j \\
		F_{ij} &  & r_{ij} < g_{ij}\  \mathrm{and}\  i \neq j\\
		0 &  & r_{ij} \geq g_{ij}\  \mathrm{and}\  i \neq j
	\end{array}\right.
.
\end{eqnarray}
Similarly, its average value can be written as
\begin{eqnarray}
\left< \Gamma_{ij} \right> & = & g_{ij} 
	\left( \frac
	{g_{ji}}
	{g_{ij}}
\right)^{\frac{1}{2}}
	\exp \left[ \frac{1}{2} \left( E_j\left( \theta \right)-E_i\left(  \theta \right) \right)\right] \\
& = & \left( g_{ij} g_{ji} \right)^{\frac{1}{2}} 
	\exp \left[ \frac{1}{2} \left( E_j\left( \theta \right)-E_i\left(  \theta \right) \right)\right]
.
\end{eqnarray}

So, we can use any connectivity scheme $\mathbf g$ in learning.  We just need to scale the non-zero, non-diagonal entries in $\mathbf \Gamma$ by $\left( \frac
	{g_{ji}}
	{g_{ij}}
\right)^{\frac{1}{2}}$ so as to compensate for the biases introduced by the connectivity scheme.

The full MPF objective function in this case is
\begin{eqnarray}
K
	& = &
\sum_{j \in \mathcal{D}} \sum_{i \notin \mathcal{D}}
	g_{ij}
	\left( 
		\frac{g_{ji}}{g_{ij}}
	\right)^\frac{1}{2}
	\exp \left[ \frac{1}{2} \left( E_j-E_i \right)\right]
\label{eq:K_finalform}
\end{eqnarray}
where the inner sum is found by averaging over samples from $g_{ij}$.

\section{Additional information on Ising spin glass example from main text}
\renewcommand{\theequation}{E-\arabic{equation}}
 % redefine the command that creates the equation no.
 \setcounter{equation}{0}  % reset counter
\renewcommand{\thefigure}{E-\arabic{figure}}
 \setcounter{figure}{0}  % reset fig counter

\subsection{Competing techniques}

The four competing techniques against which MPF was compared are: mean field theory (MFT) with Thouless-Anderson-Palmer (TAP) corrections \cite{tap_spin}, one-step and ten-step contrastive divergence \cite{Hinton02} 
(CD-1 and CD-10), and pseudolikelihood \cite{besag}.  Table~\ref{table:error_table} shows the relative performance at convergence for each technique in terms of convergence time and mean square error in coupling strengths and pairwise correlations.

MFT with TAP involves approximating 
the Gibbs free energy of the model with a second-order Plefka expansion \cite{plefka}.  MFT+TAP is fast because it involves only an inversion of the magnetic susceptibility matrix, but it can perform poorly, for instance near criticality~\cite{hertz_spin_glass}.

Contrastive divergence approximates the term involving the partition function in $\partial_\theta D_{KL}\left(\mathbf{p^{(0)}} \doublebar \mathbf{p^{(\infty)}}\left(\theta\right)\right)$, via a Markov chain which is initialized at the data distribution $\mathbf p^{(0)}$, and then truncated after only a small number of sampling steps.  It is commonly used in machine learning, and provides an effective and fast stochastic parameter update rule for learning in many probabilistic models.  However, it is not guaranteed to converge to a fixed point, and it does not correspond exactly to an objective function. The relationship between our technique and contrastive divergence is discussed in the Supplemental Material.

Pseudolikelihood approximates the joint probability distribution of a collection of random variables 
with a product of conditional distributions, where each factor is the distribution of a single random variable conditioned on the others:
\begin{align}
p(x_1,x_2,\dots,x_d) \rightarrow \prod_{i=1}^d p(x_i | x_1, \dots x_{i-1}, x_{i+1},\dots,x_d)
\end{align}
This approach often leads to surprisingly good estimates, despite the extreme nature of the approximation.
\begin{table}[t]
\caption{Mean square error in recovered coupling strengths ($\epsilon_{J}$), mean square error in pairwise correlations ($\epsilon_{\mathrm{corr}}$) and learning time for MPF versus mean field theory with TAP correction (MFT+TAP), 1-step and 10-step contrastive divergence (CD-1 and CD-10), and pseudolikelihood (PL).}
\label{table:error_table}
\vskip 0.15in
\begin{center}
\begin{small}
\begin{sc}
\begin{tabular}{lrrr}
\hline
\hline
%\abovespace\belowspace
Technique & $\epsilon_J$ & $\epsilon_{\mathrm{corr}}$& Time (s) \\
\hline
%\abovespace
MPF	&\ \ 0.0172&\ \ 0.0025& $\sim$60 \\
MFT+TAP &\ \ 7.7704&\ \ 0.0983& \ \ 0.1\\
CD-1	&\ \ 0.3196&\ \ 0.0127& $\sim$20000 \\
CD-10	&\ \ 0.3341&\ \ 0.0123& $\sim$20000\\
PL	&\ \ 0.0582&\ \ 0.0036& $\sim$800\\
\hline
\hline
%\belowspace
%\hline
\end{tabular}
\end{sc}
\end{small}
\end{center}
\vskip -0.1in
\end{table}

\subsection{Optimization steps taken for parameter estimation algorithms}
\subsubsection{Minimum Probability Flow and Pseudolikelihood}\label{LBFGSMPF}

Both Minimum Probability Flow and Pseudolikelihood have well defined objective functions and gradients.  Parameter estimation was thus performed by applying an off the shelf L-BFGS (quasi-Newton gradient descent) implementation \cite{schmidt} to their objective functions evaluated over the full training dataset $\mathcal D$.

\subsection{Contrastive Divergence}
The CD update rule was computed using the full training dataset.
The learning rate was annealed in a linear fashion from 3.0 to 0.1 to accelerate convergence.

\subsubsection{Mean Field Theory}

Mean field theory requires the computation of the inverse of the magnetic susceptibility matrix, which, for strong correlations, was often singular. A regularized pseudoinverse was used in the following manner:
\begin{align}
A = (\chi^T\chi + \lambda I)^+\chi^T,
\end{align}
where $I$ is the identity matrix, $M^+$ denotes the Moore-Penrose pseudoinverse of a matrix $M$, $\chi$ is the magnetic susceptibility $\chi_{ij} = \avg{x_ix_j}-\avg{x_i}\avg{x_j}$, and $\lambda$ is a regularizing parameter.  This technique is known as stochastic robust approximation \cite{boyd2004convex}.

\subsection{Dependence of computation time on sample size}

\begin{figure}
\center{
%\framebox[0.8\textwidth]
\parbox[c]{\linewidth}{
\center{
{\em (a)}\includegraphics[width= 0.9 \linewidth]{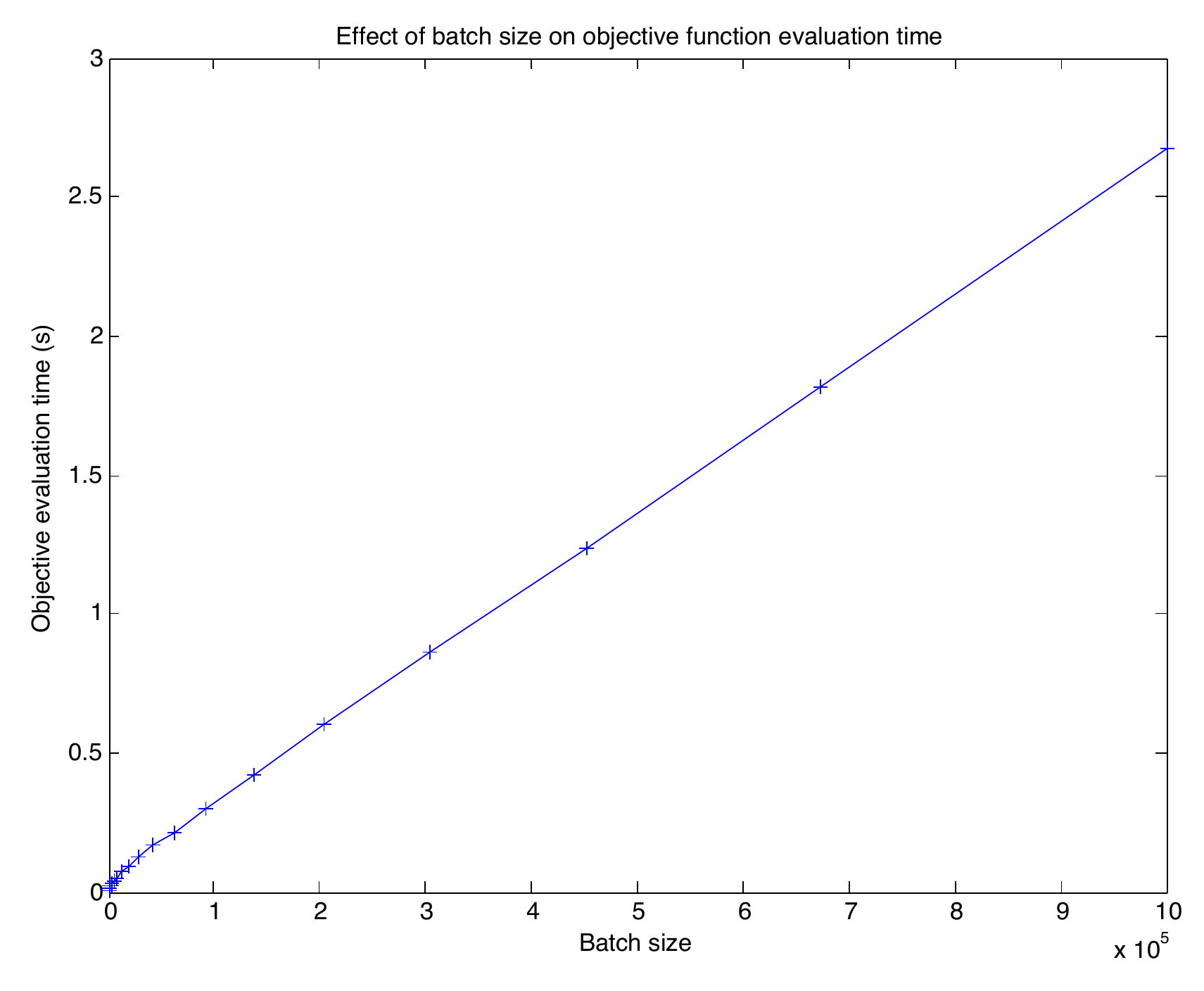} \\
{\em (b)}\includegraphics[width= 0.9 \linewidth]{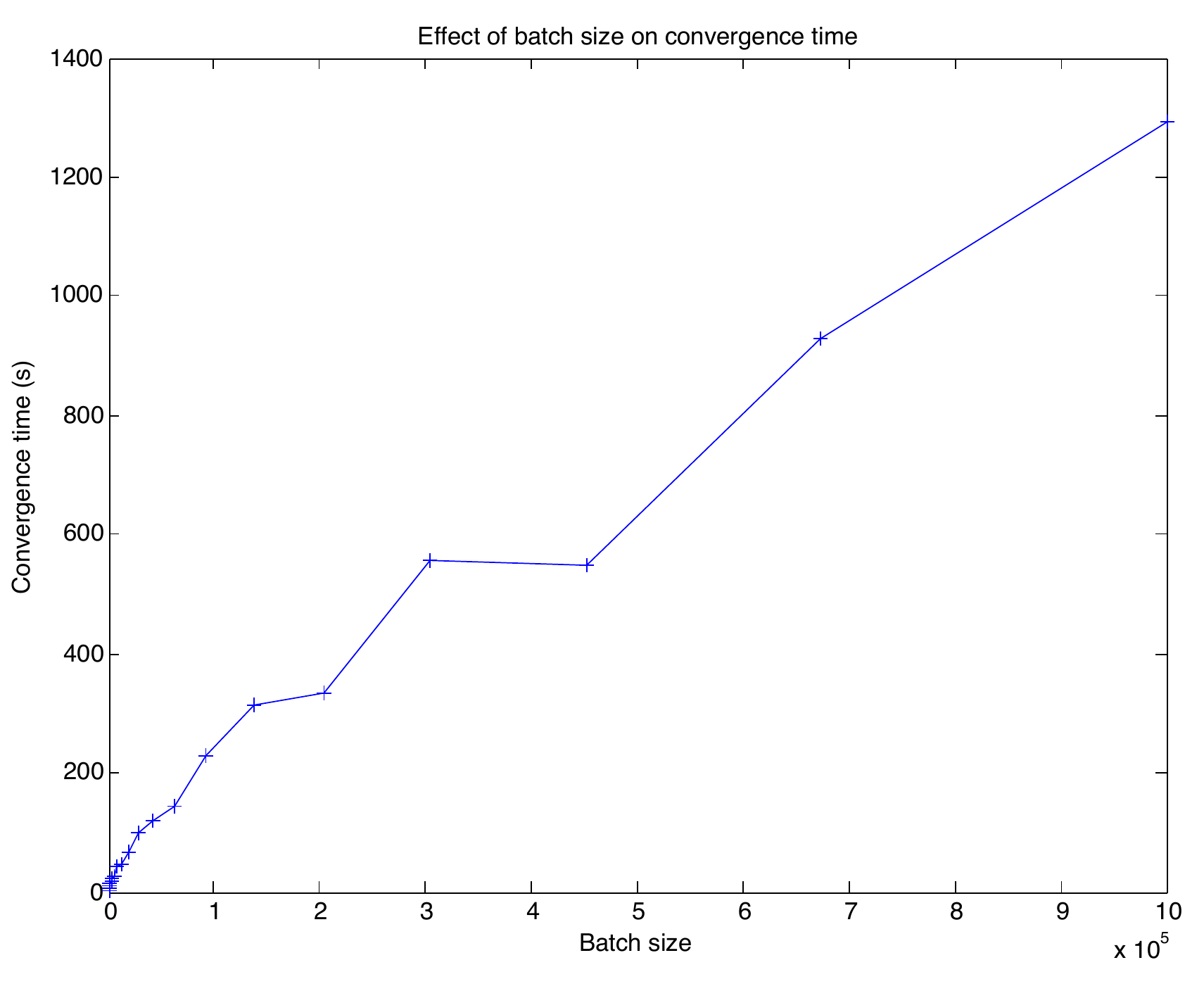}
}
}
%}
}
\caption{
\label{timefig}
The time taken for {\em (a)} evaluation of the MPF objective function and {\em (b)} convergence of the L-BFGS parameter estimation algorithm as a function of training batch size for the Ising spin glass model presented in the text.   Parameter estimation involves many parameter update steps, each of which requires reevaluating the MPF objective function and gradient.
}
\label{fig:ising time}
\end{figure}

For the Ising spin glass example described above and in the text, we measured both the time to evaluate the objective function and the time for the L-BFGS implementation in Section~\ref{LBFGSMPF} to converge as a function of batch size.  As can be seen in Figure~\ref{timefig}, for large batch size the objective function evaluation time is linear, and the convergence time is approximately linear.

\section{Additional Ising spin glass comparison}
\renewcommand{\theequation}{F-\arabic{equation}}
 % redefine the command that creates the equation no.
 \setcounter{equation}{0}  % reset counter
\renewcommand{\thefigure}{F-\arabic{figure}}
 \setcounter{figure}{0}  % reset fig counter

The Ising model has a long and storied history in physics \cite{RevModPhys.39.883} and machine learning \cite{Ackley85} and it has recently been found to be a surprisingly useful model for networks of neurons in the retina \cite{Schneidman_Nature_2006,Shlens_JN_2006}.  The ability to fit Ising models to the activity of large groups of simultaneously recorded neurons is of current interest given the increasing number of these types of data sets from the retina, cortex and other brain structures.

We fit an Ising model (fully visible Boltzmann machine) of the form
\begin{equation}
p^{(\infty)}(\mathbf{x};\mathbf{J}) = \frac{1}{Z(\mathbf{J})}\exp\left[ -\sum_{i,j} J_{ij} x_i x_j \right]
\end{equation}
to a set of $N$ $d$-element iid data samples $\left\{x^{(i)} | i = 1...N\right\}$ generated via Gibbs sampling from an Ising model as described below, where each of the $d$ elements of $\mathbf{x}$ is either 0 or 1.
Because each $x_i\in\set{0,1}$, $x_i^2=x_i$, we can write the energy function as
\begin{equation}
E(\mathbf{x};\mathbf{J}) = \sum_{i, j\neq i} J_{ij} x_i x_j + \sum_{i} J_{ii} x_i.
\end{equation}

The probability flow matrix $\mathbf{\Gamma}$ has $2^N \times 2^N$ elements, but for learning we populate it extremely sparsely,
setting
\begin{align}
\label{eqn:gamma symmetric}
	g_{ij} = g_{ji} = & 
	\left\{\begin{array}{ccc}
		1 &  & \mathrm{states\ }i\mathrm{\ and\ }j\mathrm{\ differ\ by\ single\ bit\ flip} \\
		0 &  & \mathrm{otherwise}
	\end{array}\right.
.
\end{align} 

Figure~\ref{fig:ising time} shows the average error in predicted
correlations as a function of learning time for 20,000 samples from a 40 unit, fully connected Ising model.  The final absolute correlation error is 0.0058.  The $J_{ij}$ used were graciously provided by Broderick and coauthors, and were identical to those used for synthetic data generation in the 2008 paper ``Faster solutions of the inverse pairwise Ising problem" \cite{Broderick:2007p2761}.  Training was performed on 20,000 samples so as to match the number of samples used in section III.A. of Broderick et al.  Note that given sufficient samples, the minimum probability flow algorithm would converge exactly to the right answer, as learning in the Ising model is convex (see Appendix \ref{app:convex}), and has its global minimum at the true solution.  On an 8 core 2.33 GHz Intel Xeon,
the learning converges in about $15$ seconds. Broderick et al. perform a similar learning task on a 100-CPU grid computing
cluster, with a convergence time of approximately $200$ seconds.

\begin{figure}
\center{
%\framebox[0.8\textwidth]
\parbox[c]{\linewidth}{
\center{
\includegraphics[width= 0.8 \linewidth]{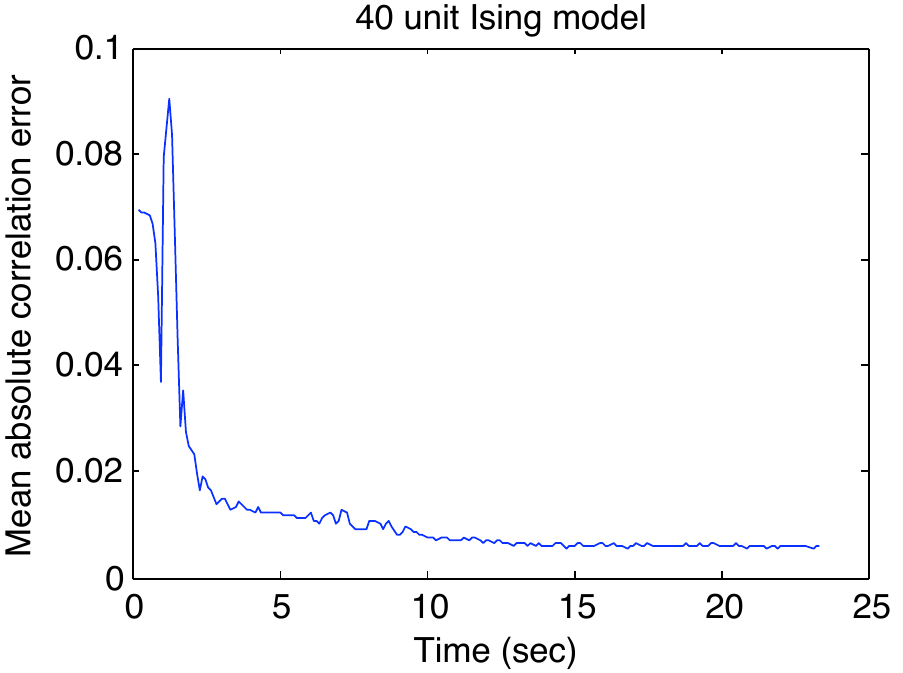}
}
}
%}
}
\caption{
A demonstration of rapid fitting of the Ising model by minimum probability flow learning.  The mean absolute error in the learned model's correlation matrix is shown as a functions of learning time for a 40 unit fully connected Ising model.  Convergence is reached in about $15$ seconds for $20,000$ samples.
}
\label{fig:ising time}
\end{figure}

\section{Continuous state space independent component analysis (ICA) \cite{ICA} model}
\renewcommand{\theequation}{G-\arabic{equation}}
 % redefine the command that creates the equation no.
 \setcounter{equation}{0}  % reset counter

Training was performed on 100,000 $10 \times 10$ pixel whitened natural image patches from the van Hateren database \cite{hateren_schaaf_1998}.  Minimization was performed by alternating between minimization of the objective function in Equation \ref{eq:K_finalform} and updates to the continuous state space connectivity function $g\left( \mb x_j, \mb x_i\right)$, as described in more detail in Section \ref{HMPF}.  Both training techniques were initialized with identical isotropic Gaussian noise (with variance $0.01$, such that each receptive field was initialized to nearly unit length), and trained on the same image patches, which accounts for the similarity of individual filters found by the algorithms.

\section{Continuous state space learning with the connectivity function set via Hamiltonian Monte Carlo} \label{HMPF}
\renewcommand{\theequation}{H-\arabic{equation}}
 % redefine the command that creates the equation no.
 \setcounter{equation}{0}  % reset counter

Choosing the connectivity matrix $g_{ij}$ for Minimum Probability Flow Learning is relatively straightforward in systems with binary or discrete state spaces.  Nearly any nearest neighbor style scheme seems to work quite well.  In continuous state spaces $\mb q \in \mathbb R^d$ however, connectivity functions $g\left( \mb q_i, \mb q_j \right)$ based on nearest neighbors prove insufficient.  For instance, if the non-zero entries in $g\left( \mb q_i, \mb q_j \right)$ are drawn from an isotropic Gaussian centered on $\mb q_j$, then several hundred non-zero $g\left( \mb q_i, \mb q_j \right)$ are required for every value of $\mb q_j$ in order to achieve effective parameter estimation in some fairly standard problems, such as receptive field estimation in Independent Component Analysis \cite{ICA}.

Qualitatively, we desire to connect every data state $\mb q_j \in \mathcal D$ to the non data states $\mb q_i$ which will be most informative for learning.  The most informative states are those which have high probability under the model distribution $p^{(\infty)}\left( \mb q \right)$.
We therefore propose to populate $g\left( \mb q_i, \mb q_j \right)$ using a Markov transition function for the model distribution.  Borrowing techniques from Hamiltonian Monte Carlo \cite{Neal:HMC} we use Hamiltonian dynamics in our transition function, so as to effectively explore the state space.

\subsection{Extending the state space}

In order to implement Hamiltonian dynamics, we first extend the state space to include auxiliary momentum variables.

The initial data and model distributions are $p^{(0)}\left( \mb q \right)$ and
\begin{align}
p^{(\infty)}\left( \mb q; \theta \right) &= \frac{\exp\left( -E\left( \mb q; \theta \right) \right)}{Z\left( \theta \right)}
.
\end{align}
with state space $\mb q \in \mathbb R^d$.  We introduce auxiliary momentum variables $\mb v \in \mathbb R^d$ for each state variable $\mb q$, and call the extended state space including the momentum variables $\mb x = \left\{ \mb q, \mb v \right\}$.  The momentum variables are given an isotropic gaussian distribution,
\begin{align}
p\left( \mb v \right) &= \frac{\exp\left( -\frac{1}{2}\mb v^T \mb v \right)}{\sqrt{2\pi}}
,
\end{align}
and the extended data and model distributions become
\begin{align}
p^{(0)}\left( \mb x \right) &=
	p^{(0)}\left( \mb q \right)
	 p\left( \mb v \right) \\
	&= 
	p^{(0)}\left( \mb q \right)
	\frac{\exp\left( -\frac{1}{2}\mb v^T \mb v \right)}{\sqrt{2\pi}} 
	\\
p^{(\infty)}\left( \mb x; \theta \right) &=
	p^{(\infty)}\left( \mb q; \theta \right) p\left( \mb v \right)
	 \\
&=
	\frac{\exp\left( -E\left( \mb q; \theta \right) \right)}{Z\left( \theta \right)}
	\frac{\exp\left( -\frac{1}{2}\mb v^T \mb v \right)}{\sqrt{2\pi}} \\
&=
	\frac{\exp\left( -H\left( \mb x; \theta \right) \right)}{Z\left( \theta \right)\sqrt{2\pi}} \\
H\left( \mb x; \theta \right) & = 
	E\left( \mb q; \theta \right) + \frac{1}{2}\mb v^T \mb v
.
\end{align}
The initial (data) distribution over the joint space $\mb x$ can be realized by drawing a momentum $\mb v$ from a uniform Gaussian distribution for every observation $\mb q$ in the dataset $\mathcal D$.

\subsection{Defining the connectivity function $g\left( \mb x_i, \mb x_j \right)$} \label{sec cont}

We connect every state $\mb x_j$ to all states which satisfy one of the following 2 criteria,
\begin{enumerate}
\item All states which share the same position $\mb q_j$, with a quadratic falloff in $g\left( \mb x_i, \mb x_j \right)$ with the momentum difference $\mb v_i - \mb v_j$.
\item The state which is reached by simulating Hamiltonian dynamics for a fixed time $t$ on the system described by $H\left( \mb x; \theta_H \right)$, and then negating the momentum.  Note that the parameter vector $\theta_H$ is used only for the Hamiltonian dynamics.
\end{enumerate}

More formally,
\begin{align}
g\left( \mb x_i, \mb x_j \right) &= \delta\left( \mb q_i - \mb q_j \right)\exp\left(-\left|\left| \mb v_i - \mb v_j \right| \right|_2^2\right) \nonumber \\
&\qquad + \delta\left( \mb x_i - \ham{\mb x_j}{\theta_H} \right) \label{eq connect}
\end{align}
where if $\mb x' = \ham{\mb x}{\theta_H}$, then $\mb x'$ is the state that results from integrating Hamiltonian dynamics for a time $t$ and then negating the momentum.  Because of the momentum negation, $\mb x = \ham{\mb x'}{\theta_H}$, and $g\left( \mb x_i, \mb x_j \right) = g\left( \mb x_j, \mb x_i \right)$.

\subsection{Discretizing Hamiltonian dynamics}

It is generally impossible to \textbf{exactly} simulate the Hamiltonian dynamics for the system described by $H\left( \mb x; \theta_H \right)$.  However, if $\ham{\mb x}{\theta_H}$ is set to simulate Hamiltonian dynamics via a series of leapfrog steps, it retains the important properties of reversibility and phase space volume conservation, and can be used in the connectivity function $g\left( \mb x_i, \mb x_j \right)$ in Equation \ref{eq connect}.
In practice, therefore, $\ham{\mb x}{\theta_H}$ involves the simulation of Hamiltonian dynamics by a series of leapfrog steps.

\subsection{MPF objective function}

The MPF objective function for continuous state spaces and a list of observations $\mathcal D$ is
\begin{align}
K\left( \theta ; \mathcal D, \theta_H \right) & = \sum_{\mb x_j \in \mathcal D} \int g\left( \mb x_i, \mb x_j \right) \nonumber \\ &\qquad
\exp\left( \frac{1}{2} \left[   
	H\left( \mb x_j; \theta \right) - H\left( \mb x_i; \theta \right) 
\right] \right)d\mb x_i
.
\end{align}
For the connectivity function $g\left( \mb x_i, \mb x_j \right)$ given in Section \ref{sec cont}, this reduces to
\begin{align}
&K\left( \theta; \mathcal D, \theta_H  \right) = \nonumber \\
&\qquad  \sum_{\mb x_j \in \mathcal D} \int \exp\left(-\left|\left| \mb v_i - \mb v_j \right| \right|_2^2\right) \nonumber \\ &\qquad\qquad\qquad
		\exp\left( \frac{1}{2} \left[   
			 \frac{1}{2}\mb v_j^T \mb v_j - -\frac{1}{2}\mb v_i^T \mb v_i
		\right] \right) d\mb v_i
\nonumber \\
&\qquad   +  \sum_{\mb x_j \in \mathcal D}  \exp\left( \frac{1}{2} \left[   
	H\left( \mb x_j; \theta \right) - H\left( \ham{\mb x_j}{\theta_H}; \theta \right) 
\right] \right)
.
\end{align}
Note that the first term does not depend on the parameters $\theta$, and is thus just a constant offset which can be ignored during optimization.  Therefore, we can say
\begin{align}
&K\left( \theta; \mathcal D, \theta_H  \right) \sim \nonumber \\ &\qquad \sum_{\mb x_j \in \mathcal D}  \exp\left( \frac{1}{2} \left[   
	H\left( \mb x_j; \theta \right) - H\left( \ham{\mb x_j}{\theta_H}; \theta \right) 
\right] \right)
.
\end{align}

Parameter estimation is performed by finding the parameter vector $\hat\theta$ which minimizes the objective function $K\left( \theta; \mathcal D, \theta_H  \right)$,
\begin{align}
\hat\theta &= \argmin_\theta K\left( \theta; \mathcal D, \theta_H  \right) 
.
\end{align}

\subsection{Iteratively improving the objective function}

The more similar $\theta_H$ is to $\theta$, the more informative $g\left( \mb x_i, \mb x_j \right)$ is for learning.  If $\theta_H$ and $\theta$ are dissimilar, then many more data samples will be required in $\mathcal D$ to effectively learn.  Therefore, we iterate the following procedure, which alternates between finding the $\hat\theta$ which minimizes $K\left( \theta; \mathcal D, \theta_H  \right)$, and improving $\theta_H$ by setting it to $\hat\theta$,
\begin{enumerate}
\item Set $\hat\theta^{t+1} = \argmin_\theta K\left( \theta; \mathcal D, \theta^t_H  \right)$
\item Set $\theta^{t+1}_H = \hat\theta^{t+1}$
\end{enumerate}
$\hat\theta^{t}$ then represents a steadily improving estimate for the parameter values which best fit the model distribution $p^{(\infty)}\left( \mb q; \theta \right)$ to the data distribution $p^{(0)}\left( \mb q \right)$, described by observations $\mathcal D$.  Practically, step 1 above will frequently be truncated early, perhaps after 10 or 100 L-BFGS gradient descent steps.

%\bibliography{prl}        % prl.bib is the name of our database

%\end{document}

\end{document}